\documentclass[a4paper,fleqn]{cas-sc}

\usepackage[numbers]{natbib}
\usepackage{bm}
\usepackage{tabularx}
\usepackage{booktabs}
\usepackage[linesnumbered,ruled,vlined]{algorithm2e}
\usepackage{graphicx}
\usepackage{caption}    % subfigure
\usepackage{subcaption} % subfigure
\usepackage{adjustbox}
\usepackage{float,lscape}
\usepackage{float}
\usepackage[leftcaption]{sidecap}
\usepackage[labelfont=bf]{caption}
\usepackage{placeins}

\def\tsc#1{\csdef{#1}{\textsc{\lowercase{#1}}\xspace}}
\tsc{WGM}
\tsc{QE}
\tsc{EP}
\tsc{PMS}
\tsc{BEC}
\tsc{DE}
\begin{document}
\let\WriteBookmarks\relax
\def\floatpagepagefraction{1}
\def\textpagefraction{.001}

%\shorttitle{Leveraging social media news}

%\shortauthors{CV Radhakrishnan et~al.}

\title [mode = title]{A Fitness-assignment Method for Evolutionary Constrained Multi-objective Optimization}                      

\author[1]{Oladayo S. Ajani}
%\cormark[1]
\ead{oladayosolomon@gmail.com}

\credit{Conceptualization, Methodology, Software, Writing - Original draft preparation, Writing - Review \& Editing}

\affiliation[1]{organization={Department of Artificial Intelligence, Kyungpook National University},
    country={South Korea}}

\author[2]{Sri Srinivasa Raju M}
\ead{msrisrinivasaraju@gmail.com}
\credit{validation, Writing - Review \& Editing}

\affiliation[2]{organization={Department of Mathematics, National Institute of Technology Silchar, }, country={India}}

\author[3]{Anand Paul}
\ead{apaul4@lsuhsc.edu}
\credit{Writing - Review \& Editing}

\affiliation[3]{{Biostatistics and Data Science Department, LSU Health Sciences Center, New Orleans, 70112, LA}, country={USA}}

% Fifth author
\author%
[1]{Rammohan Mallipeddi}
\cormark[2]
\ead{mallipeddi.ram@gmail.com}
\credit{Supervision, Resources, Validation, Writing - Review \& Editing}
\cortext[cor2]{Principal corresponding author}

\begin{abstract}
The effectiveness of Constrained Multi-Objective Evolutionary Algorithms (CMOEAs) depends on their ability to reach the different feasible regions during evolution, by exploiting the information present in infeasible solutions, in addition to optimizing the several conflicting objectives. Over the years, researchers have proposed several CMOEAs to handle  Constrained Multi-objective Optimization Problems (CMOPs). However, most of the proposed CMOEAs with scalable performance are too complex because they are either multi-staged or multi-population-based algorithms. Consequently, to ensure the simplicity of CMOEAs, researchers have proposed different fitness-assignment-based CMOEAs by combining different fitness-assignment-based methods used to solve unconstrained multi-objective problems with information regarding the feasibility of each solution. The main performance drawback of such methods is that it is difficult to design a fitness assignment method that can account for constraint violation in addition to convergence and diversity. Hence in this paper, we propose an effective single-population fitness assignment-based CMOEA referred to as $I^{c}_{SDE^+}$ that can explore different feasible regions in the search space. $I^{c}_{SDE^+}$ is a fitness assignment-based algorithm, that is an efficient fusion of constraint violation (c), \textcolor{black}{Shift-based Density Estimation (SDE)} and sum of objectives ($+$). The performance of  $I^{c}_{SDE^+}$ is favorably compared against 9 state-of-the-art CMOEAs on 6 different benchmark suites with diverse characteristics.
\end{abstract}

% \includegraphics{figs/grabs.pdf}

% % Research highlights
% \begin{highlights}
% \item Research highlights item 1
% \item Research highlights item 2
% \item Research highlights item 3
% \end{highlights}

% Keywords

\maketitle

\section{Introduction}

Optimization problems characterized by conflicting objectives and constraints are referred to as Constrained Multi-objective Optimization Problems (CMOPs) \cite{Kumar2021ABO} as in (\ref{eq:cmop})

\begin{equation}
\label{eq:cmop}
\left\{
\begin{matrix}
\text{Minimize:} \quad \bm{f}(\bm{x}) =\left [ f_{1}(\bm{x}), f_{2}(\bm{x}), ..., f_{m}(\bm{x}) \right ] \\ \\
\text{subject to:} \quad
\begin{array}{l}
g_{i}(\bm{x})  \leq  0, \quad i = 1,...,r \\ 
h_{i}(\bm{x}) = 0, \quad i = r+1,...,q \\ 
\bm{x} = (x_{1},...,x_{n}) \in \Omega  \subset \mathbb{R}^{n}
\end{array}
\end{matrix} \right. 
\end{equation}
where $\bm{f}(x) \in \mathbb{R}^{m}$ is an objective vector with $m$ objective values,  $\Omega$ denotes the variable space, $r$ and $(q - r)$ are the number of inequality and equality constraints, respectively. Employing a small positive value $\epsilon$ $(= 10^{-6})$ \cite{CUATE2020100619}, 
$(q - r)$ equality constraints can be transformed into inequality constraints (\ref{eq:2}). 
Then, the overall Constraint Violation (CV) of $x$ is given by (\ref{eq:3}).

\begin{equation}
 \label{eq:2}
g_{i}(\bm{x}) = |h_{i}(\bm{x})| - \epsilon \leq 0, \quad i = r+1, \dots, q
\end{equation}
\begin{equation}
 \label{eq:3}
CV(\bm{x})= \sum_{i=1}^{q}{\max}(0,g_{i}(\bm{x}))
\end{equation}

A solution $\bm{x}$ is termed feasible if \textcolor{black}{$CV(\bm{x}) = 0$, otherwise it is infeasible.} Given two solutions $\bm{x}_{1}$ and $\bm{x}_{2}$, $\bm{x}_{1}$ is said to dominate $\bm{x}_{2}$ (represented as $\bm{x}_{1}  \textcolor{black} {\prec} \hspace{0.05cm}  \bm{x}_{2}$) \textcolor{black}{iff $f_{j}(\bm{x}_{1}) \leq f_{j}(\bm{x}_{2})$ for each $i \in {j, . . . , m}$} and $f(\bm{x} _{1}) \neq  f(\bm{x}_{2})$. \textcolor{black}{The terms associated with a CMOP  as shown in (\ref{eq:cmop}) are briefly defined \textcolor{black}{as follows}.} 

\begin{itemize}
    \item{Unconstrained Pareto Set (UPS) and Unconstrained Pareto Front (UPF):} For any $\bm{x}^{\lozenge } \in$ UPS, $\nexists \bm{x} \in \Omega$  such that $\bm{x} \textcolor{black} {\prec} \hspace{0.05cm} \bm{x}^{\lozenge}$. UPF$ = \left \{f(\bm{x})\mid\bm{x} \in UPS\right \}$.
    \item{Constrained Pareto Set (CPS) and Constrained Pareto Front (CPF):} For any $x^{\bigstar} \in $ CPS, $\nexists \bm{x} \in \Omega$ such that $CV(\bm{x}) = 0$ and $\bm{x} \textcolor{black} {\prec} \hspace{0.05cm} \bm{x} ^{\bigstar}$. CPF $= \left \{f(\bm{x}) \mid \bm{x} \in CPS\right \}$.
\end{itemize}

As shown in Fig. 1, constraints complicate the search process \cite{ensemble} by\textcolor{black}{:} a) introducing discontinuities in search space, and/or b) altering relative arrangement of UPF and CPF. Therefore, the search process should cross multiple infeasible barriers (Fig. 1(a \& b)) or navigate through multiple narrow feasible regions (Fig. 1(c)) to reach the continuous and discrete CPF, respectively. In addition, UPF and CPF can be separated (Fig. 1(a)) or overlapped (Fig. 1(b)) or CPF $\subset$ UPF (Fig. 1(c)).
 
\begin{figure*}[!htb]
\centering
\includegraphics[width=0.98\textwidth]{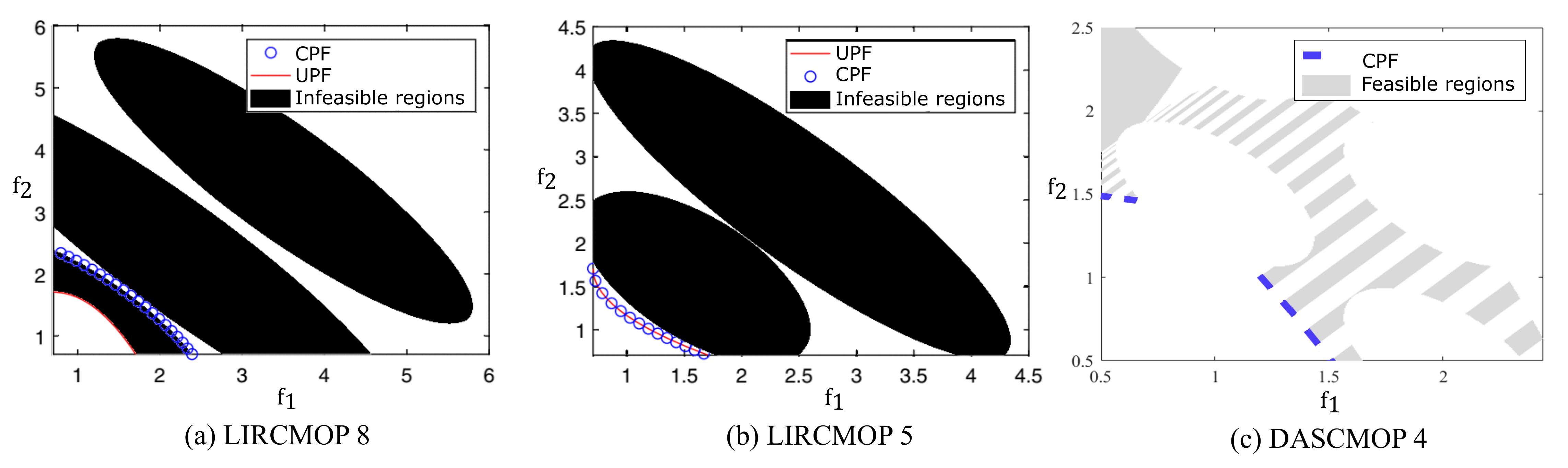}
\caption{Characteristics of CMOPs showing feasible and infeasible regions, relative location of CPF and UPF. \textcolor{black}{LIRCMOP 8 and LIRCMOP 5 are problem instances taken from LIRCMOP \cite{lircmop} while DASCMOP 4 is a problem instance taken from DASCMOP \cite{dascmop}}}
\label{fig:PF}
\end{figure*} 

\textcolor{black}{In general, depending on the mechanisms employed to resolve the underlying conflicting objectives in MOEAs in terms of selection, fitness assignments, etc, MOEAs can be broadly classified into \textcolor{black}{three main classes, namely dominance-based, decomposition-based, and indicator-based \cite{ZHOUsurvey}}. Dominance-based MOEAs also known as Pareto-based MOEAs employ the Pareto dominance principle for selection \cite{NSGA2}. The primary goal is to obtain a set of solutions whose objective value cannot be further improved on one objective without degrading on another objective. Consequently, Pareto-based MOEAs utilize the trade-off between conflicting objectives to evolve a set of non-dominated solutions. \textcolor{black}{In decomposition-based MOEAs \cite{WANG2decompose}, the multiobjective optimization problem (MOP) is converted or decomposed into a set of single-objective sub-problems that are solved collaboratively \cite{LI201952}}. Indicator-based MOEAs \cite{Indicatorsurvey} address MOPs by leveraging indicators to guide the search process and evaluate solution quality. \textcolor{black}{In other words, indicators that can assess the quality of solution sets, and consequently the contribution of a solution within a set to the overall quality} in the multiobjective space are used as selection criteria to drive convergence and diversity simultaneously.}

\textcolor{black}{Over the years, Constrained Multi-Objective Evolutionary Algorithms (CMOEAs) have shown significant potential in solving CMOPs \cite{armoea, two_arx, cmoead, ccmo, moeaddae, PPS, tige, yew-soon, yongwang}. In general, the effectiveness of CMOEAs depend on their ability to reach the different feasible regions during evolution, by exploiting the information present in infeasible solutions. \textcolor{black}{In other words, the effectiveness of CMOEAs greatly depends on the employed Constraint Handling Technique (CHT) that helps CMOEA to exploit the infeasible solutions and overcome premature convergence (Fig. \ref{fig:PF}(a \& b)) or the inability to cover the entire CPF (Fig. \ref{fig:PF}(c)). Therefore, various CHTs \cite{CoelloCoello2002TheoreticalAN,srank,superior,sadaptive,ep} have been proposed in the literature to \textcolor{black}{implement} efficient CMOEAs.} Depending on featured methodologies employed to exploit the infeasible solutions,  CMOEAs are classified as\textcolor{black}{:} fitness-based, ranking-based, multi-staged and multi-population based CMOEAs \cite{RAJU2022557,ccmo}.} 

\textcolor{black}{In fitness-based CMOEAs, solutions are assigned fitness values that reflect their level of both feasibility and convergence}. In the simplest form of fitness-based CMOEAs \cite{fitness-based,fitness-based2}, a CMOP is transformed into an unconstrained problem by adding a penalty term (reflecting the solutions' feasibility) to the actual objective values. In ranking-based CMOEAs, the objectives and the constraints are handled independently using the Constrained Dominance Principle (CDP) or combining CDP with shift-based density estimation \cite{zhou-jin} or stochastic ranking \cite{S_rank}. Although fitness-based and ranking-based  CMOEAs are simple, their performance heavily depends on the associated hyperparameters, \textcolor{black}{which if they are not appropriately set results} in severe performance degradation especially in CMOPs \textcolor{black}{characterized} by discontinuities shown in Fig. 1. \textcolor{black}{This is because such CMOPs require that some level of priority be given to infeasible solutions during evolution to overcome premature convergence.} To deal with such CMOPs, multi-staged and multi-population-based CMOEAs are gaining prominence recently. Multi-staged CMOEAs \cite{PPS,moeaddae} generally divide the evolution process into different stages to \textcolor{black}{achieve the} required balance between objectives and constraints. In other words, some stages are designed to focus on the exploration of feasible regions while others focus on enforcing diversity and uniformity in the solution set.  For example, in PPS \cite{PPS} the population is first pushed towards UPF by neglecting CV and pulled towards CPF by considering CV. Similarly, MOEADDAE \cite{moeaddae} proposed a mechanism to detect stagnation based on feasibility ratio and rate of change of CV. Once stagnation is detected, weights associated with CV are modified to assist the search in jumping out of local optima. The main drawback of multi-staged CMOEAs is the uncertainty associated with the transition between stages. In multi-population based CMOEAs \cite{multi-pop}, multiple populations co-evolve by sharing information. \textcolor{black}{ However, they often fail to achieve consistency in terms of convergence and diversity in the main population and therefore require very sophisticated mechanisms to drive such consistency \citep{multi-pop,Shi2023ANT}}. In CTAEA \cite{two_arx},  two archives are considered where one archive accounts for diversity while the other accounts for convergence. Similarly, in CCMO \cite{ccmo}, the first population solves CMOP (\ref{eq:cmop}) and the second population optimizes the objectives neglecting the constraints. Note that the diversity archive of CTAEA, second population of CCMO, and the push stage of PPS serve similar purpose of locating UPF. However, the associated mechanisms of multi-staged and multi-population based CMOEAs waste function evaluations through unnecessary exploration of infeasible regions \cite{moeaddae}.

\textcolor{black}{From the above, it can be observed that there is a need for advanced mechanisms that retain the simplicity of fitness-based methods with improved performance. Hence a class of fitness-based methods that works based on the principle of fitness assignment has been proposed where a single \cite{armoea} or multiple values \cite{tige} sometimes referred to as indicators are used to estimate the fitness of contribution of each solution in the population.  In \textcolor{black}{AR-MOEA} \cite{armoea}, \textcolor{black}{the Inverted Generation Distance (IGD) fitness assignment method \cite{IGD}} was employed, where the associated set of reference points are adapted according to the contribution of candidate solutions in an external archive. \textcolor{black}{The reference point adaptation depends on the characteristics of  each CMOP} and can seriously degrade the CMOEA performance. In TIGE\_2 \cite{tige}, \textcolor{black}{three} separate fitness assignment methods, one each for convergence, diversity and CV are combined through different ranking and balancing schemes to obtain an appropriate balance between \textcolor{black}{three} goals. \textcolor{black}{In \cite{yongwang}, \emph{m}-dimensional objective vectors corresponding} to solutions of CMOP (\ref{eq:cmop}) are transformed into scalar values by employing fitness assignment methods \textcolor{black}{such as $I_{{\epsilon}^+}$ and hypervolume (HV) contribution} developed to solve Unconstrained Multi-objective Optimization Problems (UMOPs). By combining scalar values with CV, CMOP (\ref{eq:cmop}) can be solved as a single-objective constraint problem using respective constraint handling methods such as superiority of feasible and epsilon constraint. However, the framework fails to provide diverse set of solutions corresponding to CPF. In ICMA \cite{yew-soon}, a fitness-assignment method was proposed where a cost value-based distance was introduced into objective space, which is then combined with CV to evaluate the contribution of each individual in exploring the promising areas. However, as the fitness assignment method fails to enforce diversity by itself, weight vectors are employed to divide the objective space into multiple regions. In addition, an archive is employed to store feasible solutions produced during the evolution and is pruned using environmental selection of \cite{prea}. The problem with these advanced fitness assignment-based methods is that it is difficult to design an assignment method that can account for CV in addition to convergence and diversity.}
%%%%%%%%%%%%%%

\textcolor{black}{Motivated by the need for a simple yet efficient fitness assignment-based algorithm for Constrained Evolutionary Multiobjective Optimization (CEMO), this work proposes $I^{c}_{SDE^+}$ which is an amalgamation of CV based on superiority of feasible, Sum of OBjectives (SOB) and Shift-based Density Estimation (SDE).  Efficient fusion of the \textcolor{black}{three} components enables $I^{c}_{SDE^+}$ to achieve feasible solutions with better convergence and diversity. In addition, $I^{c}_{SDE^+}$ is an efficient single-population framework, an alternative to complex multi-staged and/or multi-population approaches, to handle CMOPs. Unlike some fitness-assignment methods specifically designed for CMOPs \cite{yew-soon}, $I^{c}_{SDE^+}$ is an extension of $I_{SDE^+}$ developed to solve UMOPs.}

The rest of this paper is structured as follows. In Section II,  some background for the current work is presented. The effectiveness of \textcolor{black}{proposed $I^{c}_{SDE^+}$  algorithm} are detailed in Section III. In Section IV, the effectiveness of the proposed framework is experimentally demonstrated and compared with 9 state-of-the-art CMOEAs on various benchmark problems. Finally, Section V concludes the paper.

\section{Background on Basic Components of $I^{c}_{SDE^+}$}
\subsection{Sum of Objectives (SOB)}
Fitness assignment \cite{fitnessA} involves the transformation of \textcolor{black}{multi-objective space} into a scalar one to \textcolor{black}{facilitate an easy comparison of solutions}. It has been demonstrated that Weighted Sum of OBjectives (WSOB) \textcolor{black}{possesses} better convergence characteristics compared to other methods such as maximum ranking, favor relation and average ranking \cite{fitnessA,Qu2010,ajani2024multi}. SOB is a special case of WSOB where the weights are set to 1 as in (\ref{eqn:SB}). 
\begin{equation}
 \footnotesize
   SOB(\textcolor{black}{\bm{x}}) = \sum_{j=1}^{m}f_{j}(\textcolor{black}{\bm{x}})
 \label{eqn:SB}
\end{equation}
\textcolor{black}{To ensure that the SOB values are range-independent, normalization of the objectives is important. \textcolor{black}{Consequently}, the objectives are normalized using the maximum and minimum objective values of the current population as also employed in $I_{SDE^{+}}$ \cite{isde}.}

\subsection{Shift-based Density Estimation (SDE)}
In multi-objective optimization, SDE \cite{sde} evaluates the density of a solution (\emph{$\bm{x}$}) in population (\emph{P}) of size \emph{N} by shifting the location of the other solutions in \emph{P}, in the objective space. As SDE involves shifting solutions in the objective space, \textcolor{black}{it possesses} certain ability to impart convergence. Therefore, an \textcolor{black}{fitness-assignment-based algorithm referred to as $I_{SDE}$} given by (\ref{eqn:ISDE}) was proposed \cite{sr}.

\begin{equation}
    \textcolor{black}{I_{SDE}(\bm{x}) = \underset{\bm{y}\in P, \bm{x} \neq \bm{y} }{\textcolor{black}{\min}} \left \|f(\bm{x}) - \hat{f}({\bm{y}})\right \|}
    \label{eqn:ISDE}
\end{equation}
where $\left \|f(\bm{x}) - \hat{f}({\bm{y}})\right \|$ is the Euclidean distance. The shifted location \textcolor{black}{$\hat{f}({\bm{y}})$ of objective vector $f({\bm{y}})$ } w.r.t to $\textcolor{black}{f({\bm{x}})}$ is defined as 

\begin{equation}
   \textcolor{black}{ \textcolor{black}{\hat{f}_{j}(\bm{y})} = \left\{\begin{matrix}
\textcolor{black}{f_{j}(\bm{x}}) & \text{\textcolor{black}{if}} \quad  f_{j}(\bm{y})< \textcolor{black}{f_{j}(\bm{x}})\\ 
 f_{j}(\bm{y})& \text{\textcolor{black}{otherwise}}
\end{matrix}\right.
j \in \textcolor{black}{\{1,..., m\}}}
\label{eqn:sbde}
\end{equation}

To further enhance convergence, SOB is fused into $I_{SDE}$ to propose $I_{SDE^+}$ \cite{isde}, an effective \textcolor{black}{fitness-assignment-based algorithm for UMOPs}.

\begin{equation}
\textcolor{black}{I_{SDE^{+}}(\bm{x}) =  \underset{\bm{y}\in P_{SOB}(\bm{x})}{\textcolor{black}{\min}} \left \| f(\bm{x}) - \hat{f}({\bm{y}}) \right \|}
\label{eqn:isde}
\end{equation}
\textcolor{black}{where $P_{SOB}(\bm{x}) \subseteq {P}$ and $\bm{y}\in {P_{SOB}(\bm{x})}$ such that $ SOB(\bm{y})< SOB(\bm{x})$.}

\section{Proposed Fitness-based CMOEA ( $I^{c}_{SDE^+}$)}
\subsection{Fitness Assignment in the $I^{c}_{SDE^+}$ Framework}

\textcolor{black}{In the proposed CMOEA framework}, the purpose of the \textcolor{black}{fitness assignment} is to assist the selection of $N$ individuals from a set of $2N$ individuals that \textcolor{black}{promotes} - a) convergence, b) diversity and \textcolor{black}{c) possesses the ability to explore the infeasible barriers.} In other words, the fitness assignment should possess the ability to prioritize infeasible solutions that help CMOEA navigate the infeasible barriers while maintaining an appropriate balance between convergence and diversity among the solutions.

In CMOEAs, the focus is to first seek feasible solutions, followed by convergence and diversity. Therefore, to evaluate solutions in $I^{c}_{SDE^+}$, the population is sorted on CV, followed by SOB. In other words, feasible solutions are prioritized followed by solutions with smaller SOB. The least ranked solution is assigned a fitness value of one, the highest possible value. Then, to evaluate the fitness of a given solution \emph{p}, only the solutions in \emph{P} that are ranked lower compared to it are shifted (\ref{eqn:cisde})

\begin{equation}
   \textcolor{black}{ I^{c}_{SDE^+}(\textcolor{black}{\bm{x}}) =  \underset{\bm{y}\in P_{SC}(\bm{x}) }{\textcolor{black}{\min}} \left \| f(\bm{x}) - \hat{f}({\bm{y}}) \right \|}
    \label{eqn:cisde}
\end{equation}

\textcolor{black}{where $P_{SC}(\bm{x}) \subseteq P$ and $\bm{y} \in P_{SC}(\bm{x})$ such that }

\begin{equation}
\left\{\begin{matrix}
CV(\textcolor{black}{\bm{y}})< CV(\textcolor{black}{\bm{x}})\\ 
SOB(\textcolor{black}{\bm{y}})< SOB(\textcolor{black}{\bm{x}})  \quad \text{\textcolor{black}{if}}  \quad CV(\textcolor{black}{\bm{y}}) = CV(\textcolor{black}{\bm{x}})
\end{matrix}\right.
\label{eqn:isdec*}
\end{equation}

By setting CV of solutions to zero, which is the case in UMOPs,   (\ref{eqn:cisde})  degenerates into (\ref{eqn:isde}) and $I^{c}_{SDE^+}$ can be viewed as a generalized version of $I_{SDE^+}$ \cite{isde}.

\begin{figure}[h]
	%\vspace{-4pt}
	\centering
	\includegraphics[width=0.6\columnwidth]{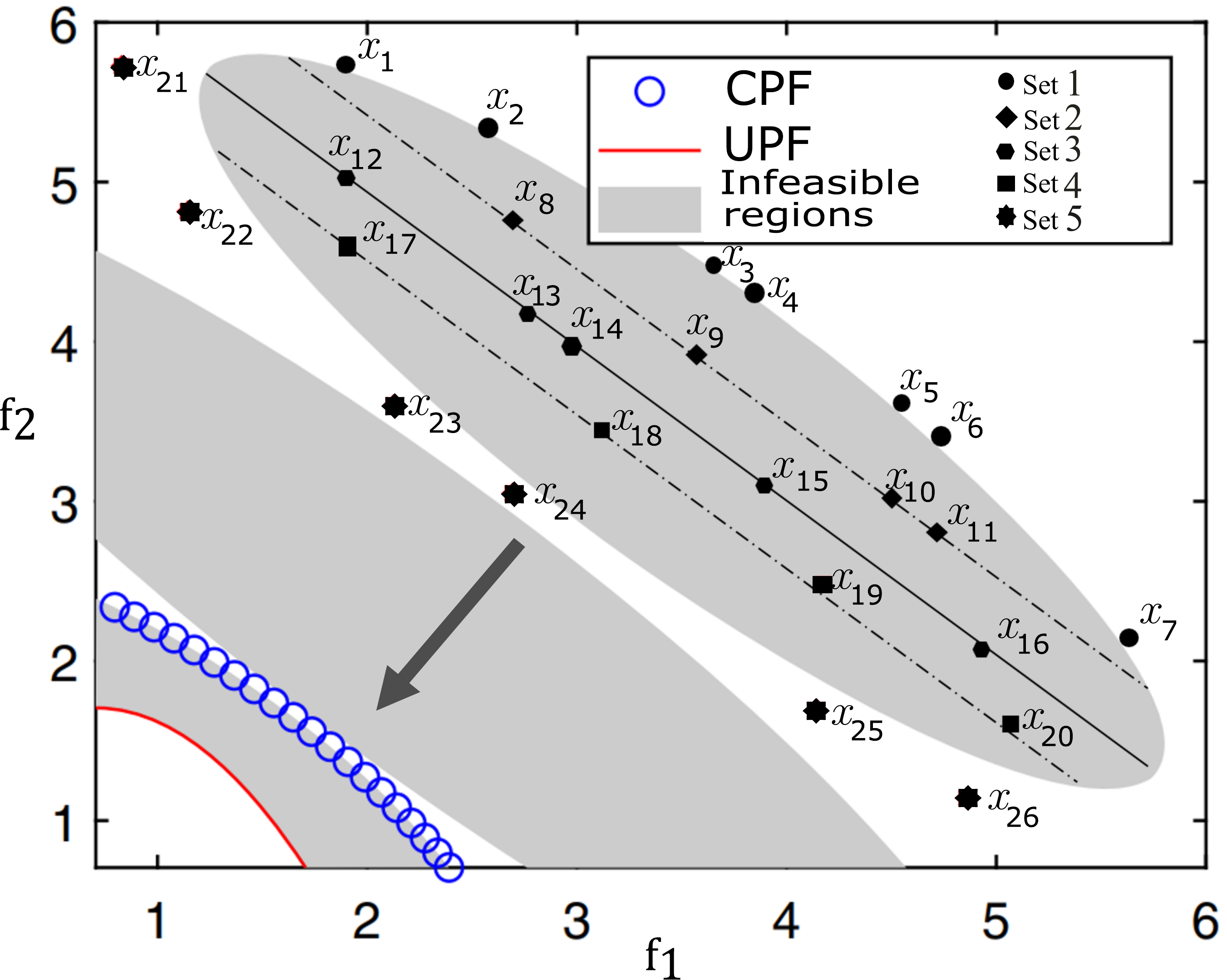}
	\caption{ \textcolor{black}{Objective space of LIRCMOP 8 \cite{lircmop} with 26 solutions divided into \textcolor{black}{five} sets and \textcolor{black}{three} stages to demonstrate the working principle of CMOEA with $I^{c}_{SDE^+}$. $CV_{\textcolor{black}{Set }3} > CV_{\textcolor{black}{Set }4}=CV_{\textcolor{black}{Set }2} > CV_{\textcolor{black}{Set }1}=CV_{\textcolor{black}{Set }5} =0 $. $SOB_{\textcolor{black}{Set }5} < SOB_{\textcolor{black}{Set }4} < SOB_{\textcolor{black}{Set }3} < SOB_{\textcolor{black}{Set }2}  < SOB_{\textcolor{black}{Set }1}$ }}
	\label{fig:con1}
	\vspace{-8pt}
\end{figure}
\begin{table}[h]
  \centering
  \caption{To demonstrate the working principle of $I^{c}_{SDE^+}$, a group of 26 solutions plotted on the objective space of LIRCMOP 8 are categorized into different sets and stages. Each ordered pair represents the rank and \textcolor{black}{ $I^{c}_{SDE^+}$-based fitness values of corresponding solutions, respectively}. Selected solutions in each stage are highlighted}
    %\begin{adjustbox}{width=0.4\columnwidth}
       \begin{tabular}{ccccc}
    \toprule
    \multicolumn{2}{c}{Solutions} & \textcolor{black}{Stage }1 & \textcolor{black}{Stage }2 & \textcolor{black}{Stage }3 \\
    \midrule
    \multirow{7}[2]{*}{\textcolor{black}{Set }1} & $\bm{x}_1$    & \textbf{[1, 1.000]} & \textbf{[1, 1.000]} & [7, 0.000] \\
          & $\bm{x}_2$    & [7, 0.040] & \textbf{[7, 0.0375]} & [13, 0.000] \\
          & $\bm{x}_3$    & [6, 0.027] & [6, 0.0270] & [12, 0.000] \\
          & $\bm{x}_4$    & [5, 0.068] & \textbf{[5, 0.0676]} & [11, 0.000] \\
          & $\bm{x}_5$    & [4, 0.027] & [4, 0.0270] & [10, 0.000] \\
          & $\bm{x}_6$    & \textbf{[2, 0.3108]} & \textbf{[2, 0.2875]} & [8, 0.000] \\
          & $\bm{x}_7$    & \textbf{[3, 0.1351]} & \textbf{[3, 0.1250]} & [9, 0.000] \\
    \midrule
    \multirow{4}[2]{*}{\textcolor{black}{Set }2} & $x_8$    & \textbf{[9, 0.122]} & [13, 0.000] & [19, 0.000] \\
          & $\bm{x}_9$    & \textbf{[11, 0.079]} & [15, 0.000] & [21, 0.000] \\
          & $\bm{x}_{10}$   & [10, 0.014] & [14, 0.000] & [20, 0.000] \\
          & $\bm{x}_{11}$   & \textbf{[8, 0.098]} & [12, 0.000] & [18, 0.000] \\
    \midrule
    \multirow{5}[2]{*}{\textcolor{black}{Set }3} & $x_{12}$   & \textbf{[13, 0.108]} & [17, 0.014] & [23, 0.000] \\
          & $\bm{x}_{13}$   & [14, 0.014] & [18, 0.014] & [24, 0.000] \\
          & $\bm{x}_{14}$   & \textbf{[12, 0.081]} & \textbf{[16, 0.041]} & [22, 0.000] \\
          & $\bm{x}_{15}$   & [15, 0.054] & [19, 0.013] & [25, 0.000] \\
          & $\bm{x}_{16}$   & \textcolor{black}{[16, 0.068]} & [20, 0.027] & [26, 0.000] \\
    \midrule
    \multirow{4}[2]{*}{\textcolor{black}{Set }4} & $x_{17}$   & -     & \textbf{[8, 0.150]} & \textbf{[14, 0.019]} \\
          & $\bm{x}_{18}$   & -     & \textbf{[9, 0.163]} & [15, 0.000] \\
          & $\bm{x}_{19}$   & -     & \textbf{[10, 0.075]} & \textbf{[16, 0.009]} \\
          & $\bm{x}_{20}$   & -     & \textbf{[11, 0.103]} & [17, 0.000] \\
    \midrule
    \multirow{6}[2]{*}{\textcolor{black}{Set }5} & $x_{21}$   & -     & -     & \textbf{[6, 0.094]} \\
          & $\bm{x}_{22}$   & -     & -     & \textbf{[3, 0.085]} \\
          & $\bm{x}_{23}$   & -     & -     & \textbf{[1, 1.000]} \\
          & $\bm{x}_{24}$   & -     & -     & \textbf{[2, 0.085]} \\
          & $\bm{x}_{25}$   & -     & -     & \textbf{[4, 0.117]} \\
          & $\bm{x}_{26}$   & -     & -     & \textbf{[5, 0.075]} \\
    \bottomrule
    \end{tabular}%
    %\end{adjustbox}
  \label{tab:addlabel}%
\end{table}%
\subsection{Working Principle of $I^{c}_{SDE^+}$}
In Fig. 2, objective space of LIRCMOP8 with 26 solutions is considered to explain the working principle \textcolor{black}{of the proposed $I^{c}_{SDE^+}$ framework and demonstrate its ability to navigate infeasible barriers (in grey)}. CMOEAs are expected to navigate (in the direction of arrow) through multiple infeasible regions, to reach the optimal CPF. To illustrate different scenarios during evolution, we consider \textcolor{black}{three} stages where solutions are divided into \textcolor{black}{five} sets as in Table I. \textcolor{black}{Stage }1, \textcolor{black}{Stage }2 and \textcolor{black}{Stage }3 contain 16, 20 and 26 solutions, respectively. In each stage, solutions are sorted first on CV followed by SOB (\ref{eqn:isdec*}) and \textcolor{black}{ $I^{c}_{SDE^+}$-based fitness values are calculated (\ref{eqn:cisde})}. In each stage, the ordered pair next to the solutions indicate the rank (\ref{eqn:isdec*}) and \textcolor{black}{ $I^{c}_{SDE^+}$-based fitness value (\ref{eqn:cisde})}, respectively. In each stage, selected solutions are highlighted (in bold). According to \cite{moeaddae}, CV across an infeasible barrier first increases to a certain point (indicated by solid line) and then decreases. Therefore, solutions on the dotted contour lines in Fig. 2 have similar CV. In other words, $CV_{\textcolor{black}{Set }3} > CV_{\textcolor{black}{Set }4}=CV_{\textcolor{black}{Set }2} > CV_{\textcolor{black}{Set }1}=CV_{\textcolor{black}{Set }5} =0 $. However, SOB decreases continuously in the direction of the arrow, i.e., $SOB_{\textcolor{black}{Set }5} < SOB_{\textcolor{black}{Set }4} < SOB_{\textcolor{black}{Set }3} < SOB_{\textcolor{black}{Set }2}  < SOB_{\textcolor{black}{Set }1}$. 

In \textcolor{black}{Stage }1, according to (\ref{eqn:isdec*}), solutions in \textcolor{black}{Set }1 are ranked ahead of \textcolor{black}{Set }2 and \textcolor{black}{Set }2 ahead of \textcolor{black}{Set }3 because $CV_{\textcolor{black}{Set }1} < CV_{\textcolor{black}{Set }2} <  CV_{\textcolor{black}{Set }3} $. In Stage1, 8 out of 16 solutions selected based on \textcolor{black}{ $I^{c}_{SDE^+}$-based fitness values} (in bold) are prioritized from different layers (Fig. 2) or sets (Table I) with a scope for exploration. In other words, even when a feasible region is found and the population contains enough feasible solutions (entire \textcolor{black}{Set }1), preference is given to infeasible solutions (from \textcolor{black}{Set }2 and \textcolor{black}{Set }3). In other words, there is a certain portion of the population that keeps exploring the potential regions, in addition to preserving the feasible solutions found. This is because when sorted on CV, feasible solutions with large objective values (\textcolor{black}{Set }1) are ranked ahead of infeasible solutions with smaller objectives (\textcolor{black}{Set }2 and \textcolor{black}{Set }3). According to (\ref{eqn:cisde}), solutions of \textcolor{black}{Set }1, when projected onto \textcolor{black}{Set }2, does not effect \textcolor{black}{ $I^{c}_{SDE^+}$-based fitness values} of solutions in \textcolor{black}{Set }2. Similarly, \textcolor{black}{Set }2 projected onto \textcolor{black}{Set }3, \textcolor{black}{does not effect }\textcolor{black}{ $I^{c}_{SDE^+}$-based fitness values} of \textcolor{black}{Set }3. In other words, \textcolor{black}{ $I^{c}_{SDE^+}$-based fitness value} of a solution is \textcolor{black}{affected} by the solutions within the same set or having similar CV. Therefore, diverse solutions from across different sets are selected.

In \textcolor{black}{Stage }2, \textcolor{black}{4} more solutions (\textcolor{black}{Set }4) are added below the solid line (Fig. 2). In \textcolor{black}{Stage }2, from the 10 of 20 solutions selected, it can be observed that no solutions from \textcolor{black}{Set }2 are selected while only one solution from \textcolor{black}{Set }3 is selected. This is because 
$CV_{\textcolor{black}{Set }4} = CV_{\textcolor{black}{Set }2} < CV_{\textcolor{black}{Set 3}}$ and $SOB_{\textcolor{black}{Set }4} < SOB_{\textcolor{black}{Set }3} < SOB_{\textcolor{black}{Set }2}$.
Therefore, solutions in \textcolor{black}{Set }4 are ranked ahead of \textcolor{black}{Set }2 and \textcolor{black}{Set }3. Because $SOB_{\textcolor{black}{Set }4} < SOB_{\textcolor{black}{Set }3} < SOB_{\textcolor{black}{Set }2}$, solutions of \textcolor{black}{Set }4 when projected on solutions of \textcolor{black}{Set }2 and \textcolor{black}{Set }3, significantly decrease $I^{c}_{SDE^+}$ values of \textcolor{black}{Set }2 and \textcolor{black}{Set }3. However, solutions in \textcolor{black}{Set }1 are ranked ahead of \textcolor{black}{Set }4 because $CV_{\textcolor{black}{Set }1} < CV_{\textcolor{black}{Set }4}$ but 
the do not \textcolor{black}{affect} the \textcolor{black}{ $I^{c}_{SDE^+}$-based fitness} values of \textcolor{black}{Set }4 because $SOB_{\textcolor{black}{Set }4} < SOB_{\textcolor{black}{Set }1}$. Therefore, solutions from \textcolor{black}{Set }1 and \textcolor{black}{Set }4 are given preference. In other words, some of the feasible solutions obtained till that point of evolution (\textcolor{black}{Set }1) are preserved while certain part of the population contains solutions (\textcolor{black}{Set }4) that helps in exploration.

In \textcolor{black}{Stage }3, 6 feasible solutions in between the infeasible barriers (\textcolor{black}{Set }5) are added (Fig. 2). Solutions in \textcolor{black}{Set }5 have smaller CV and smaller objective values than \textcolor{black}{Set }1, \textcolor{black}{Set }2, \textcolor{black}{Set }3 and \textcolor{black}{Set }4. To select 13 out 26 solutions, 8 highlighted solutions with non-zero \textcolor{black}{ $I^{c}_{SDE^+}$-based fitness values} are selected first, and the remaining solutions are randomly picked. Therefore, once the feasible solutions with better SOB exist, then evolution moves forward as the feasible solutions with better SOB replace the feasible solutions with larger SOB.

Therefore, \textcolor{black}{ $I^{c}_{SDE^+}$-based fitness} enables the CMOEA to explore the infeasible regions even when the population contains sufficient number of feasible solutions. In other words, the efficient fusion of three basic components of $I^{c}_{SDE^+}$ - CV, SOB, and SDE,  enable them complement each other in balancing both exploration and exploitation within a single population framework.  

On the contrary, certain portion of the population in the proposed framework keeps exploring the infeasible region even after reaching the optimal CPF (Fig. 1(a)), as is the case with most CMOEAs in the literature. Therefore, the number of feasible solutions in the population at any given time would be less the desired number, $\emph{N}$. However, once the search reaches the optimal CPF, over the iterations as the population evolves, different combinations of diverse sets are produced on the optimal CPF as well as in the infeasible regions below. In other words, \textcolor{black}{the  $I^{c}_{SDE^+}$ algorithm} tries to approximate the optimal CPF with a set of solutions that is less than $\emph{N}$.

\subsection{Framework of of the proposed $I^{c}_{SDE^+}$ algorithm}
Algorithm 1 presents the general structure of CMOEA with $I^{c}_{SDE^+}$. After the initial parameter setting (Line 1), starting with a uniformly initialized population (\emph{P}) of size \emph{N} (Line 2), $I^{c}_{SDE^+}$ values are evaluated (Line 3). Until a predefined stopping criterion is met, operations such as mating selection, variation, fitness evaluation and environmental selection are iterated (Lines 4-9). Finally, the population (\emph{P}) is returned.

\begin{algorithm}[h]
\footnotesize
\SetAlgoLined
\textbf{Input}: $N$ 

$\bm{P} \gets \textit{\textbf{InitializePopulation}} (N)$\\
$\bm{I^{c}_{SDE^{+}}} \gets \textit{\textbf{FitnessEvaluation}} (P)$\\

\While{not done}{
$\bm{P^{'}} \gets \textit{\textbf{MatingSelection}} (\bm{P}, N, I^{c}_{SDE^{+}})$\\
$\bm{Q} \gets \bm{P} \cup \textit{\textbf{Variation}} (\bm{P^{'}}, N)$\\
$\bm{I^{c}_{SDE^{+}}}  \gets \textit{\textbf{FitnessEvaluation}} (\bm{Q})$\\
$[\bm{P, I^{c}_{SDE^{+}}}]  \gets \textit{\textbf{EnvironmentalSelection}}(\bm{Q}, N, \bm{I^{c}_{SDE^{+}}} )$\\
}
\textbf{Output}: $\bm{P}$
\caption{Framework of $\bm{I^{c}_{SDE^{+}}}$}
\end{algorithm}

\begin{algorithm}[h]
\footnotesize
\SetAlgoLined
\textbf{Input}:$P , N, \bm{I^{c}_{SDE^{+}}}$ \\
$\bm{P} \gets  \phi $\\
\While{$\mid \bm{P'}\mid \hspace{2pt} < N$}{
select two individuals $x$ and $y$ randomly from $\bm{P}$\\
\uIf{$I^{c}_{SDE+}(\bm{x}) > I^{c}_{SDE+}(\bm{y})$}
{$\bm{P'} \gets \bm{P'} \cup \hspace{2pt}(\bm{x})$}
\Else{$\bm{P'} \gets \bm{P'} \cup \hspace{2pt} (\bm{y})$}
}
\textbf{Output}: $\bm{P'}$
\caption{$\bm{I^{c}_{SDE^{+}}}$ based Mating Selection}
\end{algorithm}

\begin{algorithm}[h]
\footnotesize
\SetAlgoLined
\textbf{Input}: $Q$, $N$, $\bm{I^{c}_{SDE^{+}}}$ \\
sort solutions in $Q$ in descending order of $\bm{I^{c}_{SDE^{+}}}$\\
$[\bm{P, I^{c}_{SDE^{+}}}]  \gets$ $N$ solutions with large fitness values are selected \textcolor{black}{and ties are resolved randomly}\\
\textbf{Output}: $\bm{P, I^{c}_{SDE^{+}}}$
\caption{$\bm{I^{c}_{SDE^{+}}}$ based Environmental Selection}
\end{algorithm}

As in Algorithm 2, during mating selection, based on $I^{c}_{SDE^+}$ promising solutions from the immediate population are selected through binary tournament selection. In binary tournament selection, out of the two randomly selected solutions, the solution with highest \textcolor{black}{ $I^{c}_{SDE^+}$-based fitness value} is preferred (lines 4–9). The solutions selected during mating selection are used to produce new solutions using combination of variation operators such as a) simulated binary crossover and polynomial mutation, or b) Differential Evolution (DE) operator and polynomial mutation \cite{moeaddae}. Finally, during environmental selection (Algorithm 3), $\emph{N}$ individuals with the highest \textcolor{black}{ $I^{c}_{SDE^+}$-based fitness values} are picked from the union of the current population and offspring population produced through mating and variation operators.

\section{Experimental Setup, Results and Discussion}
The performance of the proposed $I^{c}_{SDE^+}$ algorithm is evaluated on 6 different benchmark suites, namely; CF \cite{cf}, RWCMOP \cite{rwop}, LIRCMOP \cite{lircmop}, DASCMOP \cite{dascmop}, MW \cite{mw} and CDTLZ \cite{cdtlz}.  \textcolor{black}{These benchmark suites form the largest set of test suites ranging from synthetic to real-world problems often employed within the constrained EMO community to evaluate the performance of constrained EMO algorithms. We decided to perform our comparative analysis based on all these benchmarks to ensure that our analysis is not biased or a result of cherry-picked problem suites.}  \textcolor{black}{Similar to MOEADDAE \cite{moeaddae}, $I^{c}_{SDE^+}$ also employs DE operator and polynomial mutation to optimize CF, RWCMOP, LIRCMOP and DASCMOP test suites, while GA simulated binary crossover and polynomial mutation are employed to optimize MW and CDTLZ test suites.} \textcolor{black}{However, results of the proposed algorithm using DE and GA operators respectively are presented for all the featured test suites in Appendix A.}  The performance of \textcolor{black}{CMOEA with} $I^{c}_{SDE^+}$ is compared with 9 state-of-the-art CMOEAs namely - \textcolor{black}{AR-MOEA} \cite{armoea}, CTAEA \cite{two_arx}, CMOEAD\cite{cmoead}, CCMO \cite{ccmo}, MOEADDAE \cite{moeaddae}, PPS \cite{PPS}, TiGE\_2 \cite{tige}, ICMA \cite{yew-soon}, HypECDP \cite{yongwang}. CMOEAD is a traditional decomposition-based algorithm combined with the superiority of feasible CHT to solve CMOPs and can be used as a baseline. CTAEA and CCMO are multi-population CMOEAs. PPS and MOEADDAE are multi-stage CMOEAs. \textcolor{black}{AR-MOEA}, TIGE\_2, ICMA, and HypECDP are indicator-based CMOEAs. \textcolor{black}{To evaluate the performance of CMOEAs, the hypervolume (HV) performance indicator is used. The simulations are carried out in the PLATEMO \cite{PlatEMO} framework and employ the standard parameters and protocols. \textcolor{black}{PLATEMO is a Matlab platform for Evolutionary Multi-objective Optimization (EMO) usually employed for comparative analysis in the EMO community. It features several benchmark suites with different characteristics and known optimal Pareto fronts with standard reference points for metric calculation. }}

Each CMOEA is run 30 times on each problem instance. The results in terms of HV and statistical comparison based on the Wilcoxon signed rank test are presented in Tables  \ref{tab:cf} - \ref{tab:RWCMOP}.  Specifically, Tables \ref{tab:cf}, \ref{tab:cdtlz}, \ref{tab:dascmop}, \ref{tab:lircmop}, \ref{tab:mw}  and \ref{tab:RWCMOP} summarize the results on CF \cite{cf}, CDTLZ \cite{cdtlz}, DASCMOP \cite{dascmop}, LIRCMOP \cite{lircmop},  MW \cite{mw} and RWCMOP \cite{rwop}, respectively. In each table, the mean and standard deviation values of the best performing CMOEA (i.e. highest mean HV) for each problem instance are highlighted. In addition, the symbols $-$,$+$, and $=$ next to HV values indicate that the performance of $I^{c}_{SDE^+}$ is superior, inferior and similar to the corresponding CMOEA.

The performance of baseline CMOEAD is not consistent across the different test suites showing significantly lower results compared to $I^{c}_{SDE^+}$. In other words, $I^{c}_{SDE^+}$ is superior or comparable to CMOEAD on 87.01\% of the test instances. In problem instances such as CF10, LIRCMOP5 and LIRCMOP6, CMOEAD fails to find feasible solutions in all the 30 runs unlike the other CMOEAs. 

In comparison with multi-stage based CMOEAs, the performance of $I^{c}_{SDE^+}$ is superior or comparable to MOEADDAE and PPS on 70.13\% and 71.43\% of the test instances, respectively. However, the superior performance of multi-stage based CMOEAs compared to $I^{c}_{SDE^+}$ can be observed on LIRCMOP test suite where PPS and MOEADDAE perform better than $I^{c}_{SDE^+}$  on 10 test instances each. The elevated performance of multi-stage based CMOEAs on LIRCMOP test suite can be attributed to the alternating exploration and exploitation stages based on the employed switching mechanism. In most of the test instances, this process is expected to waste the function evaluations. However, the mechanism appropriately suits the characteristics of LIRCMOP test suite. 

In comparison with multi-population based CMOEAs, $I^{c}_{SDE^+}$ performs better or comparable on 77.92\% and 66.23\% of test instances. While multi-population based CMOEAs fails on most test suites, they demonstrate competitive performance of DASCMOP and MW test suites. The superior performance of multi-population based CMOEAs can be attributed to their design where one of the population always explores with the aim of finding the Unconstrained Pareto Front (UCF) while the other explores the Constrained Pareto Front (CPF). In CMOPs such as DASCMOP where in most of problem instances CPF $\subset$ UPF, the population exploring the UPF enables the CMOEA navigate the infeasible barriers. However, the use of multi-populations results in the wastage of function evaluations.   

Among the indicator-based CMOEAs, the performance of TIGE\_2 is significantly inferior compared to $I^{c}_{SDE^+}$. In other words, $I^{c}_{SDE^+}$ performs better or comparable on 92.21\% of the test instances. Compared to ARMOEA and HypECDP the performance of $I^{c}_{SDE^+}$ is better or comparable on 74.03\% and 77.92\% of the test instances. However, both ARMOEA and HypECDP demonstrate superior performance compared to $I^{c}_{SDE^+}$ on RWCMOP test suite by performing better on 14 and 16 test instances, respectively out of 25 test instances. In addition, the performance of $I^{c}_{SDE^+}$ in comparison with ICMA where a specialized indicator was developed for constrained multi-objective problems is better or comparable in 77.92\% of the test instances. The improved performance of ICMA compared to $I^{c}_{SDE^+}$ can be observed on CF test suite.

From the analysis, it is evident the performance of state-of-the-art CMOEAs is not consistent across the different test suites considered. In other words, alleviate performance on test suite compared to $I^{c}_{SDE^+}$ is compensated by degraded performance on the other test suites. However, the performance of $I^{c}_{SDE^+}$ is consistently ranked among the top three performing CMOEAs on each test suite. The superior performance of $I^{c}_{SDE^+}$ can be attributed to the complementary nature of the three components - constraint violation, sum of objectives and shift-based density estimation. In addition, $I^{c}_{SDE^+}$ enables the development of single population-based CMOEA framework facilitating the efficient use of function evaluations.

\begin{landscape}
% Table generated by Excel2LaTeX from sheet 'HV'
\begin{table*}[!h]
  \centering
 \caption{\leftskip=-0.0cm Performance Comparison of $I^{c}_{SDE^+}$ against state-of-the-art CMOEAs in terms of HV (mean and standard deviation) as well as Wilcoxon’s signed-rank test  on CF test suite }
\label{tab:cf}%
\begin{adjustbox}{width=1.3\textwidth,right}
    \begin{tabular}{ccccccccccccccc}
    \toprule
    Problem & N     & M     & D     & FEs   & ARMOEA & CTAEA & CMOEAD & CCMO  & MOEADDAE & PPS   & TiGE\_2 & ICMA  & HypECDP & cISDE \\
    \midrule
    CF1   & 100   & 2     & 10    & 300000 & 5.5577e-1 (1.35e-3) - & 5.2057e-1 (4.78e-3) - & 5.5495e-1 (3.14e-3) - & 5.6316e-1 (2.94e-4) - & 5.6454e-1 (3.04e-4) - & \textbf{5.6493e-1 (3.91e-4) =} & 4.9069e-1 (1.24e-2) - & 5.6170e-1 (1.86e-3) - & 5.5566e-1 (1.71e-3) - & 5.6477e-1 (1.51e-4) \\
    CF2   & 100   & 2     & 10    & 300000 & 6.1717e-1 (2.76e-2) - & 6.4862e-1 (1.24e-2) - & 5.7526e-1 (2.88e-2) - & 6.6141e-1 (7.73e-3) - & 6.4395e-1 (2.46e-2) - & \textbf{6.7709e-1 (1.88e-3) +} & 5.5169e-1 (4.14e-2) - & 6.7447e-1 (1.25e-3) + & 6.4796e-1 (2.18e-2) - & 6.7000e-1 (6.64e-3) \\
    CF3   & 100   & 2     & 10    & 300000 & 1.7355e-1 (5.05e-2) - & 2.0321e-1 (3.71e-2) = & 1.4995e-1 (5.21e-2) - & 2.1032e-1 (3.18e-2) = & 1.5889e-1 (5.13e-2) - & 1.8836e-1 (5.31e-2) = & 1.0272e-1 (5.62e-2) - & \textbf{2.5495e-1 (3.55e-2) +} & 1.8692e-1 (3.95e-2) - & 2.1433e-1 (3.71e-2) \\
    CF4   & 100   & 2     & 10    & 300000 & 4.2673e-1 (2.35e-2) - & 4.3068e-1 (3.12e-2) - & 3.8908e-1 (3.92e-2) - & 4.5354e-1 (2.95e-2) - & 3.7334e-1 (6.62e-2) - & 4.8584e-1 (4.27e-2) = & 3.0289e-1 (5.97e-2) - & \textbf{5.1620e-1 (3.11e-3) +} & 4.2496e-1 (3.61e-2) - & 4.6880e-1 (1.63e-2) \\
    CF5   & 100   & 2     & 10    & 300000 & 2.1344e-1 (6.65e-2) - & 2.5057e-1 (8.81e-2) - & 2.3831e-1 (7.99e-2) - & 2.7530e-1 (7.23e-2) = & 2.1248e-1 (7.30e-2) - & 3.4156e-1 (4.50e-2) + & 2.4056e-1 (7.13e-2) - & \textbf{3.6204e-1 (6.34e-2) +} & 3.0012e-1 (5.60e-2) = & 3.1160e-1 (5.34e-2) \\
    CF6   & 100   & 2     & 10    & 300000 & 6.4143e-1 (1.54e-2) - & 6.3617e-1 (1.88e-2) - & 6.0690e-1 (2.92e-2) - & 6.6355e-1 (1.14e-2) - & 6.6872e-1 (1.46e-2) - & 6.5600e-1 (1.37e-2) - & 5.7653e-1 (3.25e-2) - & \textbf{6.8619e-1 (2.76e-3) +} & 6.4935e-1 (1.49e-2) - & 6.8133e-1 (1.22e-2) \\
    CF7   & 100   & 2     & 10    & 300000 & 4.5906e-1 (6.19e-2) = & 4.4986e-1 (8.76e-2) = & 4.2541e-1 (1.03e-1) - & 4.8488e-1 (7.64e-2) = & 3.3679e-1 (1.32e-1) - & 4.4027e-1 (1.04e-1) = & 3.8496e-1 (6.87e-2) - & \textbf{6.1806e-1 (1.32e-2) +} & 4.2788e-1 (1.02e-1) - & 4.7125e-1 (1.11e-1) \\
    CF8   & 150   & 3     & 10    & 300000 & 4.0315e-1 (2.82e-2) - & 3.2177e-1 (1.58e-2) - & 3.7416e-1 (3.16e-2) - & 3.5867e-1 (1.02e-1) - & 4.2471e-1 (5.19e-2) - & 3.9209e-1 (1.78e-2) - & 8.7528e-2 (2.65e-2) - & 4.4114e-1 (1.34e-2) - & 9.8389e-2 (5.12e-2) - & 5.0122e-1 (1.97e-2) \\
    CF9   & 150   & 3     & 10    & 300000 & 4.4186e-1 (2.58e-2) - & 4.1332e-1 (1.59e-2) - & 4.2686e-1 (3.12e-2) - & 4.4742e-1 (4.24e-2) - & 4.6729e-1 (3.33e-3) - & 4.7919e-1 (1.97e-2) - & 1.2176e-1 (4.63e-2) - & 5.2011e-1 (4.04e-3) - & 2.3151e-1 (3.46e-2) - & 5.3216e-1 (1.09e-2) \\
    CF10  & 150   & 3     & 10    & 300000 & 2.1601e-1 (1.46e-1) = & 1.9397e-1 (5.70e-2) - & 0.0000e+0 (0.00e+0) - & 1.7229e-1 (4.92e-2) - & 1.6024e-1 (8.76e-2) - & 2.6441e-1 (7.12e-2) - & 9.8521e-2 (1.50e-2) - & \textbf{3.8068e-1 (4.29e-2) =} & 9.1731e-2 (0.00e+0) = & 3.4766e-1 (9.26e-2) \\
    \midrule
    \multicolumn{5}{c}{+/-/=}             & 0/8/2 & 0/8/2 & 0/10/0 & 0/7/3 & 0/10/0 & 2/4/4 & 0/10/0 & 6/3/1 & 0/8/2 &  \\
    \bottomrule
    \end{tabular}%
   \end{adjustbox}
   \vspace{-0.5cm}
\end{table*}%
% Table generated by Excel2LaTeX from sheet 'HV'
\begin{table*}[!h]
  \centering
\caption{\leftskip=-0cm Performance Comparison of $I^{c}_{SDE^+}$ against state-of-the-art CMOEAs in terms of HV (mean and standard deviation) as well as Wilcoxon’s signed-rank test  on CDTLZ test suite }
 \label{tab:cdtlz}%
\begin{adjustbox}{width=1.3\textwidth,center}
    \begin{tabular}{ccccccccccccccc}
    \toprule
    Problem & N     & M     & D     & FEs   & ARMOEA & CTAEA & CMOEAD & CCMO  & MOEADDAE & PPS   & TiGE\_2 & ICMA  & HypECDP & cISDE \\
    \midrule
    C1\_DTLZ1 & 92    & 3     & 7     & 46000 & 8.3350e-1 (6.93e-3) + & 8.3363e-1 (5.79e-3) + & 8.3443e-1 (7.30e-3) + & 8.3282e-1 (6.99e-3) + & \textbf{8.3577e-1 (6.35e-3) +} & 7.9849e-1 (2.14e-2) - & 2.7192e-1 (1.14e-1) - & 8.1799e-1 (7.41e-3) - & 2.8445e-1 (9.39e-2) - & 8.2892e-1 (4.48e-3) \\
    C1\_DTLZ3 & 92    & 3     & 12    & 92000 & 7.4927e-2 (1.81e-1) - & 4.4962e-1 (1.86e-1) - & 2.2183e-1 (2.75e-1) - & 5.5813e-1 (1.51e-3) - & 5.5601e-1 (1.60e-3) - & 4.3894e-1 (1.75e-1) - & 0.0000e+0 (0.00e+0) - & 3.4868e-1 (2.05e-1) - & 2.4047e-2 (7.38e-2) - & \textbf{5.5974e-1 (1.29e-3)} \\
    C2\_DTLZ2 & 92    & 3     & 12    & 23000 & 5.1146e-1 (1.85e-3) + & 5.0317e-1 (3.16e-3) = & 5.1474e-1 (2.56e-4) + & 5.1368e-1 (1.90e-3) + & \textbf{5.1533e-1 (1.22e-3) +} & 4.7943e-1 (1.31e-2) - & 4.0809e-1 (2.55e-2) - & 4.8224e-1 (3.83e-3) - & 4.9536e-1 (3.06e-3) - & 5.0214e-1 (3.02e-3) \\
    C3\_DTLZ1 & 92    & 3     & 7     & 69000 & 2.5251e-1 (5.95e-2) - & 2.7051e-1 (4.94e-2) - & 2.8657e-1 (6.42e-2) = & 2.9213e-1 (1.04e-2) - & 2.9909e-1 (1.21e-2) - & 2.6706e-1 (5.81e-2) - & 9.0213e-3 (1.91e-2) - & 2.5865e-1 (1.60e-2) - & 1.3504e-2 (2.68e-2) - & \textbf{3.1650e-1 (9.61e-3)} \\
    C3\_DTLZ4 & 92    & 3     & 12    & 69000 & 7.4292e-1 (1.05e-1) - & 7.8459e-1 (1.20e-3) - & 7.8651e-1 (4.88e-2) - & 7.7946e-1 (4.58e-2) - & 3.8013e-1 (1.69e-1) - & 7.5409e-1 (3.25e-2) - & 7.6175e-1 (3.72e-3) - & 7.3759e-1 (6.79e-3) - & 6.5890e-1 (1.28e-1) - & \textbf{7.9352e-1 (4.45e-4)} \\
    \midrule
    \multicolumn{5}{c}{+/-/=}   & 2/3/0 & 1/3/1 & 2/2/1 & 2/3/0 & 2/3/0 & 0/5/0 & 0/5/0 & 0/5/0 & 0/5/0 &  \\
    \bottomrule
    \end{tabular}%
    \end{adjustbox}
       \vspace{-0.5cm}
\end{table*}%
% Table generated by Excel2LaTeX from sheet 'HV'
\begin{table*}[!h]
  \centering
 \caption{ \leftskip=-0cm Performance Comparison of $I^{c}_{SDE^+}$ against state-of-the-art CMOEAs in terms of HV (mean and standard deviation) as well as Wilcoxon’s signed-rank test  on DASCMOP test suite }
 \label{tab:dascmop}%
  \begin{adjustbox}{width=1.3\textwidth,right}
    \begin{tabular}{ccccccccccccccc}
    \toprule
    Problem & N     & M     & D     & FEs   & ARMOEA & CTAEA & CMOEAD & CCMO  & MOEADDAE & PPS   & TiGE\_2 & ICMA  & HypECDP & cISDE \\
    \midrule
    DASCMOP1 & 300   & 2     & 30    & 300000 & 4.8466e-3 (4.59e-3) - & 1.6243e-1 (5.03e-4) - & 6.0020e-3 (6.55e-3) - & 6.3241e-3 (5.10e-3) - & 2.8654e-2 (4.22e-2) - & 1.9649e-1 (4.16e-3) - & 1.5082e-1 (7.99e-3) - & 1.9242e-1 (4.75e-3) - & 7.3850e-3 (4.95e-3) - & 2.0013e-1 (2.41e-3) \\
    DASCMOP2 & 300   & 2     & 30    & 300000 & 2.4770e-1 (6.80e-3) - & 3.1053e-1 (1.12e-2) - & 2.4167e-1 (9.44e-3) - & 2.5830e-1 (5.49e-3) - & 2.7170e-1 (1.80e-2) - & \textbf{3.4978e-1 (4.15e-4) +} & 3.1445e-1 (5.99e-3) - & 3.3973e-1 (2.37e-3) - & 2.4883e-1 (6.63e-3) - & 3.4872e-1 (7.39e-4) \\
    DASCMOP3 & 300   & 2     & 30    & 300000 & 2.0852e-1 (1.96e-4) - & 2.5921e-1 (9.02e-3) - & 1.9595e-1 (4.86e-2) - & 2.1746e-1 (1.53e-2) - & 2.3012e-1 (3.27e-2) - & 3.0461e-1 (1.31e-2) - & 2.5189e-1 (2.08e-2) - & 3.0999e-1 (9.16e-4) - & 2.1071e-1 (8.27e-3) - & \textbf{3.1180e-1 (1.59e-4)} \\
    DASCMOP4 & 300   & 2     & 30    & 300000 & 1.3511e-1 (5.86e-2) - & 2.0090e-1 (1.74e-3) + & \textbf{2.0390e-1 (1.79e-4) +} & 2.0244e-1 (3.21e-3) + & 2.0327e-1 (3.45e-4) + & 1.8800e-1 (1.40e-2) - & 1.8929e-1 (5.66e-3) - & 1.9634e-1 (1.25e-2) = & 1.9887e-1 (1.37e-2) + & 1.9713e-1 (1.39e-2) \\
    DASCMOP5 & 300   & 2     & 30    & 300000 & 2.8362e-1 (1.13e-1) - & 3.4951e-1 (3.87e-4) + & 3.5124e-1 (2.97e-4) + & 3.5166e-1 (1.90e-4) + & 3.5039e-1 (3.58e-4) + & 3.4257e-1 (1.11e-2) = & 3.4115e-1 (1.95e-3) - & 3.4708e-1 (2.19e-3) = & 3.4426e-1 (4.24e-2) - & 3.4767e-1 (1.17e-3) \\
    DASCMOP6 & 300   & 2     & 30    & 300000 & 1.8426e-1 (9.12e-2) - & 3.0911e-1 (8.78e-4) = & 2.6626e-1 (7.89e-2) = & 3.1080e-1 (4.84e-3) + & 2.9200e-1 (2.31e-2) = & 2.7689e-1 (3.46e-2) - & 3.0782e-1 (2.21e-3) = & 3.1020e-1 (1.06e-3) + & 2.7406e-1 (7.95e-2) - & 2.9833e-1 (2.88e-2) \\
    DASCMOP7 & 300   & 3     & 30    & 300000 & \textbf{2.9258e-1 (2.89e-4) +} & 2.8662e-1 (1.27e-3) + & 2.9177e-1 (4.98e-4) + & 2.9239e-1 (5.06e-4) + & 2.9061e-1 (4.17e-4) + & 2.6876e-1 (2.72e-2) = & 2.6469e-1 (2.87e-3) - & 2.8160e-1 (4.63e-3) + & 1.8877e-1 (2.08e-2) - & 2.7286e-1 (1.20e-2) \\
    DASCMOP8 & 300   & 3     & 30    & 300000 & \textbf{2.1319e-1 (2.75e-4) +} & 2.0498e-1 (7.36e-4) + & 2.1152e-1 (3.97e-4) + & 2.1298e-1 (5.31e-4) + & 1.9854e-1 (3.87e-3) + & 1.6437e-1 (2.95e-2) - & 1.9651e-1 (1.71e-3) + & 1.9481e-1 (5.25e-3) = & 1.1289e-1 (8.01e-3) - & 1.9236e-1 (4.78e-3) \\
    DASCMOP9 & 300   & 3     & 30    & 300000 & 1.2613e-1 (1.10e-2) - & 1.9484e-1 (7.05e-3) = & 1.6269e-1 (4.82e-2) = & 1.3619e-1 (7.36e-3) - & 1.5411e-1 (3.31e-2) - & \textbf{2.0006e-1 (1.89e-3) +} & 1.5290e-1 (2.09e-2) - & 1.5751e-1 (9.76e-3) - & 9.6044e-2 (5.15e-3) - & 1.9556e-1 (4.76e-3) \\
    \midrule
    \multicolumn{5}{c}{+/-/=}  & 2/7/0 & 4/3/2 & 4/3/2 & 5/4/0 & 4/4/1 & 2/5/2 & 1/7/1 & 2/4/3 & 1/8/0 &  \\
    \bottomrule
    \end{tabular}%
\end{adjustbox}
\end{table*}

% Table generated by Excel2LaTeX from sheet 'HV'
\begin{table*}[!h]
  \centering
\caption{\leftskip=-0cm Performance Comparison of $I^{c}_{SDE^+}$ against state-of-the-art CMOEAs in terms of HV (mean and standard deviation) as well as Wilcoxon’s signed-rank test  on LIRCMOP test suite }
    \label{tab:lircmop}%
    \begin{adjustbox}{width=1.3\textwidth,right}
    \begin{tabular}{ccccccccccccccc}
    \toprule
    Problem & N     & M     & D     & FEs   & ARMOEA & CTAEA & CMOEAD & CCMO  & MOEADDAE & PPS   & TiGE\_2 & ICMA  & HypECDP & cISDE \\
    \midrule
    LIRCMOP1 & 300   & 2     & 30    & 300000 & 1.2513e-1 (7.10e-3) - & 1.4622e-1 (2.80e-2) - & 1.1732e-1 (8.94e-3) - & 1.5035e-1 (1.67e-2) - & 2.0660e-1 (1.75e-2) - & \textbf{2.3707e-1 (8.23e-4) +} & 1.5997e-1 (7.29e-3) - & 2.1830e-1 (4.22e-3) - & 1.2142e-1 (6.30e-3) - & 2.2942e-1 (1.47e-3) \\
    LIRCMOP2 & 300   & 2     & 30    & 300000 & 2.3935e-1 (9.53e-3) - & 3.0861e-1 (1.52e-2) - & 2.4003e-1 (1.21e-2) - & 2.6328e-1 (1.74e-2) - & 3.4396e-1 (1.24e-2) = & \textbf{3.6021e-1 (5.76e-4) +} & 2.9146e-1 (7.29e-3) - & 3.4320e-1 (5.13e-3) - & 2.4085e-1 (1.13e-2) - & 3.5243e-1 (1.24e-3) \\
    LIRCMOP3 & 300   & 2     & 30    & 300000 & 1.1153e-1 (9.17e-3) - & 1.2257e-1 (2.19e-2) - & 1.0395e-1 (1.27e-2) - & 1.3651e-1 (1.29e-2) - & 1.2277e-1 (2.90e-2) - & \textbf{2.0547e-1 (1.03e-3) +} & 1.4044e-1 (8.61e-3) - & 1.9084e-1 (6.23e-3) - & 1.1087e-1 (9.86e-3) - & 1.9667e-1 (4.00e-3) \\
    LIRCMOP4 & 300   & 2     & 30    & 300000 & 2.0534e-1 (1.32e-2) - & 2.2988e-1 (3.23e-2) - & 1.9652e-1 (1.67e-2) - & 2.2951e-1 (1.73e-2) - & 2.6063e-1 (2.75e-2) - & \textbf{3.1515e-1 (5.81e-4) +} & 2.4491e-1 (1.02e-2) - & 2.9913e-1 (6.04e-3) - & 2.0725e-1 (1.29e-2) - & 3.0401e-1 (4.38e-3) \\
    LIRCMOP5 & 300   & 2     & 30    & 300000 & 0.0000e+0 (0.00e+0) - & 9.6995e-3 (3.69e-2) - & 0.0000e+0 (0.00e+0) - & 1.6395e-1 (1.94e-2) - & \textbf{2.9371e-1 (2.65e-5) +} & 2.9351e-1 (6.91e-5) + & 4.8731e-2 (6.31e-2) - & 2.7258e-1 (5.41e-2) = & 0.0000e+0 (0.00e+0) - & 2.9142e-1 (6.69e-4) \\
    LIRCMOP6 & 300   & 2     & 30    & 300000 & 0.0000e+0 (0.00e+0) - & 0.0000e+0 (0.00e+0) - & 0.0000e+0 (0.00e+0) - & 1.2147e-1 (1.23e-2) - & \textbf{1.9896e-1 (4.05e-5) +} & 1.9885e-1 (5.52e-5) + & 4.6648e-2 (4.19e-2) - & 1.9857e-1 (2.21e-4) + & 0.0000e+0 (0.00e+0) - & 1.8063e-1 (3.18e-2) \\
    LIRCMOP7 & 300   & 2     & 30    & 300000 & 2.3589e-1 (4.50e-2) - & 2.4633e-1 (1.11e-2) - & 8.1771e-2 (1.18e-1) - & 2.5332e-1 (9.44e-3) - & 2.9274e-1 (6.23e-4) + & \textbf{2.9582e-1 (3.37e-4) +} & 2.3305e-1 (1.48e-2) - & 2.5296e-1 (1.79e-2) - & 2.2544e-1 (7.70e-2) - & 2.9070e-1 (1.37e-3) \\
    LIRCMOP8 & 300   & 2     & 30    & 300000 & 1.3807e-1 (1.08e-1) - & 1.5387e-1 (9.85e-2) - & 2.9412e-2 (7.63e-2) - & 2.4184e-1 (9.92e-3) - & 2.9480e-1 (1.73e-4) + & \textbf{2.9603e-1 (7.81e-5) +} & 2.0405e-1 (1.46e-2) - & 2.6038e-1 (1.32e-2) - & 1.1311e-1 (1.15e-1) - & 2.8733e-1 (1.43e-2) \\
    LIRCMOP9 & 300   & 2     & 30    & 300000 & 1.3837e-1 (5.86e-2) - & 4.2727e-1 (2.90e-2) - & 2.1712e-1 (8.69e-2) - & 4.1246e-1 (6.28e-2) - & \textbf{5.6805e-1 (5.21e-5) +} & 4.6151e-1 (1.96e-2) - & 3.6410e-1 (1.88e-2) - & 5.0391e-1 (3.16e-2) - & 1.7040e-1 (8.85e-2) - & 5.5537e-1 (5.39e-3) \\
    LIRCMOP10 & 300   & 2     & 30    & 300000 & 1.0742e-1 (7.53e-2) - & 5.5099e-1 (6.53e-2) - & 4.0143e-1 (1.98e-1) - & 6.2571e-1 (2.05e-2) - & \textbf{7.0920e-1 (5.60e-5) +} & 6.9034e-1 (3.67e-2) - & 1.8806e-1 (2.64e-2) - & 7.0770e-1 (2.97e-4) + & 1.4066e-1 (1.37e-1) - & 7.0165e-1 (2.72e-3) \\
    LIRCMOP11 & 300   & 2     & 30    & 300000 & 2.1468e-1 (7.60e-2) - & 6.3668e-1 (1.04e-2) - & 2.6733e-1 (9.94e-2) - & 6.6293e-1 (1.59e-2) - & \textbf{6.9395e-1 (2.80e-5) +} & 6.2118e-1 (7.06e-2) = & 3.4816e-1 (5.63e-2) - & 6.7837e-1 (1.58e-2) - & 2.5140e-1 (6.63e-2) - & 6.9307e-1 (1.11e-3) \\
    LIRCMOP12 & 300   & 2     & 30    & 300000 & 3.3402e-1 (8.47e-2) - & 5.7665e-1 (8.54e-3) - & 4.2111e-1 (6.65e-2) - & 5.3951e-1 (4.52e-2) - & \textbf{6.2034e-1 (2.00e-5) +} & 5.6832e-1 (9.79e-3) - & 5.0397e-1 (1.13e-2) - & 6.1107e-1 (1.07e-2) = & 3.2430e-1 (1.08e-1) - & 6.0485e-1 (2.82e-2) \\
    LIRCMOP13 & 300   & 3     & 30    & 300000 & 5.9184e-4 (1.09e-4) - & 5.7650e-1 (7.57e-4) + & 5.5589e-4 (2.53e-5) - & \textbf{5.7746e-1 (4.48e-4) +} & 5.7356e-1 (5.43e-4) + & 5.5410e-1 (3.29e-3) + & 2.7524e-1 (8.50e-2) - & 5.7003e-1 (4.67e-4) + & 9.5660e-5 (2.17e-4) - & 5.2426e-1 (9.31e-3) \\
    LIRCMOP14 & 300   & 3     & 30    & 300000 & 1.6391e-3 (2.46e-4) - & 5.7554e-1 (8.03e-4) + & 1.5741e-3 (4.16e-5) - & \textbf{5.7571e-1 (4.06e-4) +} & 5.7434e-1 (7.15e-4) + & 5.6417e-1 (1.56e-3) + & 3.5311e-1 (7.02e-2) - & 5.7047e-1 (6.18e-4) + & 3.0104e-4 (4.48e-4) - & 5.2999e-1 (5.99e-3) \\
    \midrule
    \multicolumn{5}{c}{+/-/=}             & 0/14/0 & 2/12/0 & 0/14/0 & 2/12/0 & 10/3/1 & 10/3/1 & 0/14/0 & 4/8/2 & 0/14/0 &  \\
    \bottomrule
    \end{tabular}%
\end{adjustbox}
\end{table*}%
\end{landscape}

\begin{landscape}
% Table generated by Excel2LaTeX from sheet 'HV'
\begin{table*}[htbp]
  \centering
\caption{ \leftskip=-0cm Performance Comparison of $I^{c}_{SDE^+}$ against state-of-the-art CMOEAs in terms of HV (mean and standard deviation) as well as Wilcoxon’s signed-rank test  on MW test suite }
\label{tab:mw}
    \begin{adjustbox}{width=1.3\textwidth,right}
    \begin{tabular}{ccccccccccccccc}
    \toprule
    Problem & N     & M     & D     & FEs   & ARMOEA & CTAEA & CMOEAD & CCMO  & MOEADDAE & PPS   & TiGE\_2 & ICMA  & HypECDP & cISDE \\
    \midrule
    MW1   & 100   & 2     & 15    & 60000 & 4.4856e-1 (9.49e-2) - & 4.8884e-1 (3.10e-4) - & 4.6529e-1 (3.30e-2) - & \textbf{4.8979e-1 (2.07e-4) +} & 4.8617e-1 (8.14e-3) = & 4.2282e-1 (1.06e-1) - & 4.3400e-1 (4.88e-2) - & 4.8653e-1 (7.47e-4) - & 4.7941e-1 (2.41e-2) = & 4.8910e-1 (2.73e-4) \\
    MW2   & 100   & 2     & 15    & 60000 & 5.4100e-1 (2.58e-2) - & 5.6056e-1 (1.19e-2) = & 5.4976e-1 (1.28e-2) - & 5.5240e-1 (1.43e-2) = & 4.5154e-1 (8.37e-2) - & 3.7752e-1 (7.32e-2) - & 5.2374e-1 (2.14e-2) - & \textbf{5.6896e-1 (9.79e-3) +} & 5.4420e-1 (1.49e-2) - & 5.5905e-1 (1.23e-2) \\
    MW3   & 100   & 2     & 15    & 60000 & 5.1871e-1 (1.00e-1) = & \textbf{5.4420e-1 (5.98e-4) +} & 5.4159e-1 (5.57e-3) - & 5.4380e-1 (7.23e-4) = & 5.4237e-1 (8.24e-4) - & 5.4284e-1 (7.91e-4) - & 5.2102e-1 (2.19e-2) - & 5.3844e-1 (1.36e-3) - & 5.4058e-1 (1.78e-2) - & 5.4390e-1 (5.96e-4) \\
    MW4   & 100   & 3     & 15    & 60000 & 8.3914e-1 (1.15e-2) + & 8.3809e-1 (2.02e-4) = & 8.3935e-1 (2.14e-3) + & \textbf{8.4134e-1 (4.91e-4) +} & 8.2179e-1 (7.83e-2) - & 7.4558e-1 (1.05e-1) - & 7.9301e-1 (1.86e-2) - & 8.2437e-1 (3.27e-3) - & 5.0729e-1 (8.31e-2) - & 8.3803e-1 (1.38e-3) \\
    MW5   & 100   & 2     & 15    & 60000 & 1.8737e-1 (1.00e-1) - & 3.1313e-1 (4.88e-3) - & 2.9813e-1 (6.75e-2) = & 3.2188e-1 (7.47e-3) - & 3.0203e-1 (3.39e-2) - & 2.2070e-1 (9.99e-2) - & 2.8289e-1 (1.17e-2) - & 3.1642e-1 (2.19e-3) - & 1.9675e-1 (1.11e-1) - & \textbf{3.2302e-1 (5.48e-4)} \\
    MW6   & 100   & 2     & 15    & 60000 & 2.8218e-1 (2.62e-2) - & 3.1214e-1 (8.30e-3) = & 3.0828e-1 (1.15e-2) = & 2.9895e-1 (1.86e-2) - & 1.6075e-1 (8.83e-2) - & 9.6163e-2 (8.46e-2) - & 2.3467e-1 (6.78e-2) - & \textbf{3.1308e-1 (9.53e-3) =} & 2.7970e-1 (4.44e-2) - & 3.1217e-1 (1.37e-2) \\
    MW7   & 100   & 2     & 15    & 60000 & 3.7871e-1 (6.37e-2) - & 4.0866e-1 (8.15e-4) + & 4.1084e-1 (4.65e-4) + & \textbf{4.1213e-1 (5.01e-4) +} & 4.1198e-1 (4.57e-4) + & 4.1164e-1 (4.44e-4) + & 3.8789e-1 (4.68e-3) - & 4.0131e-1 (1.66e-3) - & 4.0023e-1 (4.24e-2) - & 4.0655e-1 (1.42e-3) \\
    MW8   & 100   & 3     & 15    & 60000 & 5.0828e-1 (6.42e-2) - & 5.2023e-1 (1.51e-2) - & 5.2644e-1 (1.45e-2) = & \textbf{5.3396e-1 (1.50e-2) =} & 4.2902e-1 (8.40e-2) - & 3.4873e-1 (9.89e-2) - & 4.7769e-1 (1.94e-2) - & 4.8027e-1 (1.19e-2) - & 3.5384e-1 (4.25e-2) - & 5.3305e-1 (1.08e-2) \\
    MW9   & 100   & 2     & 15    & 60000 & 3.1163e-1 (1.43e-1) - & 3.9147e-1 (1.88e-3) - & 3.6266e-1 (7.65e-2) - & 3.8380e-1 (7.25e-2) - & 3.3900e-1 (1.36e-1) = & 2.6690e-1 (1.56e-1) - & 3.0002e-1 (1.20e-1) - & 3.3866e-1 (1.89e-2) - & 3.6798e-1 (7.21e-2) - & \textbf{3.9431e-1 (2.89e-3)} \\
    MW10  & 100   & 2     & 15    & 60000 & 3.4800e-1 (8.63e-2) - & \textbf{4.4135e-1 (1.10e-2) +} & 3.9908e-1 (3.62e-2) = & 4.0897e-1 (2.24e-2) = & 2.9257e-1 (9.74e-2) - & 1.8890e-1 (8.87e-2) - & 3.9396e-1 (3.06e-2) - & 4.2691e-1 (1.79e-2) + & 3.4146e-1 (9.38e-2) - & 4.1623e-1 (1.90e-2) \\
    MW11  & 100   & 2     & 15    & 60000 & 2.7879e-1 (3.15e-2) - & 4.4150e-1 (1.24e-3) - & 3.6304e-1 (8.36e-2) - & 4.4594e-1 (2.68e-3) + & 4.4382e-1 (1.03e-3) - & \textbf{4.4726e-1 (2.11e-4) +} & 4.3178e-1 (2.51e-3) - & 4.2631e-1 (6.35e-3) - & 3.3158e-1 (8.26e-2) - & 4.4448e-1 (5.79e-4) \\
    MW12  & 100   & 2     & 15    & 60000 & 5.1226e-1 (1.72e-1) - & 6.0061e-1 (5.66e-4) - & 5.8204e-1 (1.03e-1) = & 5.6445e-1 (1.46e-1) - & 4.8384e-1 (2.34e-1) - & 4.3122e-1 (2.39e-1) - & 4.3403e-1 (2.45e-1) - & 5.8479e-1 (3.66e-3) - & 5.3158e-1 (1.85e-1) - & \textbf{6.0397e-1 (3.97e-4)} \\
    MW13  & 100   & 2     & 15    & 60000 & 3.5649e-1 (8.50e-2) - & \textbf{4.6086e-1 (1.03e-2) +} & 4.3085e-1 (3.87e-2) - & 4.4505e-1 (1.80e-2) = & 3.3373e-1 (9.57e-2) - & 2.9515e-1 (1.03e-1) - & 3.1744e-1 (4.16e-2) - & 4.4756e-1 (1.08e-2) = & 3.9451e-1 (7.33e-2) - & 4.5051e-1 (1.24e-2) \\
    MW14  & 100   & 3     & 15    & 60000 & 4.6785e-1 (1.03e-2) + & 4.6523e-1 (3.29e-3) = & 4.4032e-1 (3.09e-3) - & 4.7173e-1 (2.31e-3) + & \textbf{4.7246e-1 (6.98e-3) +} & 4.4052e-1 (8.19e-3) - & 4.4814e-1 (5.27e-3) - & 4.5449e-1 (3.19e-3) - & 4.4858e-1 (5.23e-3) - & 4.6430e-1 (6.80e-3) \\
    \midrule
    \multicolumn{5}{c}{+/-/=}             & 2/11/1 & 4/6/4 & 2/7/5 & 5/4/5 & 2/10/2 & 2/12/0 & 0/14/0 & 2/10/2 & 0/13/1 &  \\
    \bottomrule
    \end{tabular}%
\end{adjustbox}
\end{table*}%
% Table generated by Excel2LaTeX from sheet 'HV'
\begin{table*}[htbp]
  \centering
\caption{\leftskip=-0cm Performance Comparison of $I^{c}_{SDE^+}$ against state-of-the-art CMOEAs in terms of HV (mean and standard deviation) as well as Wilcoxon’s signed-rank test  on RWCMOP test suite }
    \label{tab:RWCMOP}%
    \begin{adjustbox}{width=1.3\textwidth,right}
    \begin{tabular}{ccccccccccccccc}
    \toprule
    Problem & N     & M     & D     & FEs   & ARMOEA & CTAEA & CMOEAD & CCMO  & MOEADDAE & PPS   & TiGE\_2 & ICMA  & HypECDP & cISDE \\
    \midrule
    RWCMOP1 & 80    & 2     & 4     & 20000 & 6.0661e-1 (7.69e-4) + & 6.0358e-1 (1.20e-3) + & 1.0893e-1 (1.04e-4) - & 6.0369e-1 (7.84e-4) + & 5.5680e-1 (1.67e-2) - & 4.8070e-1 (5.85e-2) - & 5.1628e-1 (2.32e-2) - & 9.8773e-2 (2.51e-2) - & \textbf{6.0922e-1 (1.75e-4) +} & 5.9815e-1 (1.61e-3) \\
    RWCMOP2 & 80    & 2     & 5     & 20000 & 1.8904e-1 (1.47e-1) - & 1.0888e-1 (1.52e-1) - & 2.8555e-1 (1.40e-1) + & 2.3415e-1 (1.56e-1) = & 6.2865e-4 (3.01e-3) - & 3.9137e-1 (1.01e-3) + & 1.0551e-1 (1.28e-1) - & \textbf{3.9731e-1 (2.11e-3) +} & 2.3798e-1 (1.34e-1) = & 2.7352e-1 (7.01e-2) \\
    RWCMOP3 & 80    & 2     & 3     & 20000 & 8.9781e-1 (6.69e-4) - & 8.6608e-1 (2.05e-2) - & 1.2293e-1 (3.55e-2) - & 8.9740e-1 (1.10e-3) - & \textbf{9.0094e-1 (1.80e-4) +} & 8.3937e-1 (1.57e-2) - & 7.3557e-1 (1.29e-1) - & 9.1945e-2 (1.57e-3) - & 2.5116e-1 (2.78e-1) - & 9.0001e-1 (3.79e-4) \\
    RWCMOP4 & 80    & 2     & 4     & 20000 & 8.5398e-1 (7.96e-3) + & 8.5128e-1 (8.78e-3) + & 1.3982e-2 (2.99e-2) - & 8.5261e-1 (7.47e-3) + & 8.5699e-1 (1.44e-3) + & 8.4129e-1 (1.72e-2) + & 4.5797e-1 (1.64e-1) - & 2.6353e-1 (3.97e-2) - & \textbf{8.6033e-1 (2.86e-3) +} & 8.3719e-1 (8.55e-3) \\
    RWCMOP5 & 80    & 2     & 4     & 20000 & 4.3314e-1 (1.16e-3) + & 4.2897e-1 (4.36e-3) = & 4.1822e-1 (9.29e-3) - & 4.3256e-1 (1.65e-3) + & 4.2473e-1 (4.27e-3) - & 4.3266e-1 (5.49e-4) + & 3.9801e-1 (7.39e-3) - & 3.8887e-1 (2.03e-2) - & \textbf{4.3389e-1 (1.79e-3) +} & 4.3154e-1 (9.27e-4) \\
    RWCMOP6 & 80    & 2     & 7     & 20000 & 2.7678e-1 (9.80e-5) + & 2.2018e-1 (6.18e-2) - & 2.7632e-1 (2.22e-4) + & 2.7669e-1 (3.12e-4) + & 2.7666e-1 (1.42e-4) + & 2.7664e-1 (2.14e-4) + & 2.7133e-1 (1.67e-3) - & 2.6241e-1 (2.86e-3) - & \textbf{2.7725e-1 (9.59e-5) +} & 2.7449e-1 (3.52e-4) \\
    RWCMOP7 & 80    & 2     & 4     & 20000 & 4.8351e-1 (5.48e-4) - & 4.8149e-1 (5.11e-3) - & 4.8017e-1 (7.48e-4) - & 4.8407e-1 (7.61e-5) = & 4.8302e-1 (3.68e-4) - & 4.8299e-1 (4.36e-4) - & 4.6850e-1 (1.51e-2) - & 4.7903e-1 (5.40e-3) - & \textbf{4.8457e-1 (2.70e-5) +} & 4.8403e-1 (5.21e-4) \\
    RWCMOP8 & 105   & 3     & 7     & 26250 & 2.6079e-2 (7.14e-5) + & 2.5765e-2 (1.10e-3) = & 9.6240e-3 (7.31e-4) - & 2.5926e-2 (8.41e-5) + & 2.5756e-2 (1.02e-4) - & 2.4160e-2 (8.94e-4) - & 2.0471e-2 (2.60e-4) - & 2.3757e-2 (3.94e-4) - & \textbf{2.6114e-2 (6.98e-5) +} & 2.5819e-2 (5.54e-4) \\
    RWCMOP9 & 80    & 2     & 4     & 20000 & 4.0956e-1 (8.11e-5) + & 4.0761e-1 (1.01e-3) - & 5.3069e-2 (5.07e-5) - & 4.0865e-1 (2.22e-4) - & 3.8236e-1 (4.69e-3) - & 3.8245e-1 (8.62e-3) - & 3.2232e-1 (1.26e-2) - & 1.3417e-1 (8.36e-4) - & \textbf{4.0963e-1 (5.17e-4) +} & 4.0916e-1 (2.33e-4) \\
    RWCMOP10 & 80    & 2     & 2     & 20000 & 8.4052e-1 (2.19e-3) - & 8.4091e-1 (3.70e-3) - & 7.9512e-2 (3.94e-4) - & 8.3857e-1 (3.36e-3) - & 8.4692e-1 (1.72e-4) - & 8.4682e-1 (1.89e-4) - & 8.3978e-1 (2.52e-3) - & 7.2210e-1 (8.99e-2) - & 8.4683e-1 (8.10e-4) = & \textbf{8.4729e-1 (1.14e-4)} \\
    RWCMOP11 & 212   & 5     & 3     & 53000 & 9.4658e-2 (2.49e-3) - & 9.9651e-2 (4.08e-4) - & 5.9562e-2 (1.01e-3) - & 9.7550e-2 (6.75e-4) - & 1.0102e-1 (3.57e-4) - & 1.0034e-1 (3.29e-4) - & 9.8074e-2 (8.08e-4) - & 7.9270e-2 (2.98e-3) - & 1.0042e-1 (8.31e-4) - & 1.0213e-1 (2.36e-4) \\
    RWCMOP12 & 80    & 2     & 4     & 20000 & 5.5860e-1 (1.64e-3) + & 5.4098e-1 (1.10e-2) = & 7.0200e-2 (1.81e-2) - & 5.5216e-1 (4.81e-3) + & 5.4794e-1 (6.92e-3) = & 5.3273e-1 (2.54e-2) - & 5.4799e-1 (2.43e-3) = & 0.0000e+0 (0.00e+0) - & \textbf{5.6167e-1 (2.88e-4) +} & 5.4663e-1 (6.38e-3) \\
    RWCMOP13 & 105   & 3     & 7     & 26250 & \textbf{9.0299e-2 (7.58e-5) +} & 8.6679e-2 (4.27e-3) - & 8.9835e-2 (1.80e-3) - & 8.8736e-2 (1.89e-4) - & 9.0250e-2 (6.06e-5) + & 8.9525e-2 (3.65e-4) - & 8.6741e-2 (7.11e-4) - & 8.4126e-2 (1.59e-3) - & 9.0262e-2 (7.46e-5) + & 8.9986e-2 (1.19e-4) \\
    RWCMOP14 & 80    & 2     & 5     & 20000 & 6.1681e-1 (1.27e-3) + & 6.1657e-1 (1.12e-3) + & 1.1076e-1 (2.23e-2) - & 6.1404e-1 (1.75e-3) + & 6.1037e-1 (3.93e-3) - & 5.7034e-1 (2.02e-2) - & 3.0629e-1 (1.30e-1) - & 5.1941e-1 (3.11e-2) - & \textbf{6.1863e-1 (1.05e-3) +} & 6.1238e-1 (2.19e-3) \\
    RWCMOP15 & 80    & 2     & 3     & 20000 & 5.4057e-1 (3.71e-3) + & 5.3445e-1 (1.60e-2) + & 6.6011e-2 (8.19e-6) - & 5.3768e-1 (3.88e-3) + & 3.8793e-1 (8.03e-2) - & 5.3056e-1 (5.83e-3) = & 5.0865e-1 (1.06e-2) - & 8.3703e-2 (5.63e-3) - & \textbf{5.4167e-1 (8.25e-4) +} & 5.2379e-1 (1.47e-2) \\
    RWCMOP16 & 80    & 2     & 2     & 20000 & 7.6245e-1 (1.44e-5) - & 7.6139e-1 (7.47e-4) - & 7.9117e-2 (8.58e-5) - & 7.6156e-1 (3.09e-4) - & 7.6081e-1 (2.34e-3) - & 7.6161e-1 (9.67e-4) - & 7.4203e-1 (7.16e-3) - & 8.9209e-2 (1.41e-17) - & \textbf{7.6392e-1 (1.80e-5) +} & 7.6274e-1 (5.27e-4) \\
    RWCMOP17 & 105   & 3     & 6     & 26250 & 2.6790e-1 (1.32e-2) + & 2.4680e-1 (3.08e-2) + & 2.3219e-1 (6.08e-2) = & \textbf{2.7743e-1 (3.09e-2) +} & 2.1541e-1 (6.06e-2) = & 2.2474e-1 (6.00e-2) = & 2.4858e-1 (5.48e-2) = & 2.4245e-1 (2.53e-2) = & 2.6470e-1 (6.61e-2) + & 2.3174e-1 (3.62e-2) \\
    RWCMOP18 & 80    & 2     & 3     & 20000 & 4.0507e-2 (1.39e-6) + & 4.0265e-2 (1.25e-4) - & 4.0262e-2 (2.71e-5) - & 4.0494e-2 (3.92e-6) + & 4.0465e-2 (1.17e-5) + & 4.0438e-2 (2.15e-5) + & 3.9215e-2 (5.16e-4) - & 4.0440e-2 (2.69e-5) + & \textbf{4.0509e-2 (1.00e-6) +} & 4.0365e-2 (7.61e-5) \\
    RWCMOP19 & 105   & 3     & 10    & 26250 & 3.1663e-1 (1.15e-2) - & 1.3887e-1 (6.07e-2) - & 1.8323e-1 (3.75e-2) - & 3.2135e-1 (9.09e-3) - & 2.7673e-1 (3.28e-2) - & 3.1075e-1 (2.33e-2) - & 3.1874e-1 (1.17e-2) - & 1.6750e-1 (8.86e-2) - & 3.2962e-1 (7.46e-3) - & \textbf{3.4772e-1 (4.49e-3)} \\
    RWCMOP20 & 80    & 2     & 4     & 20000 & \textbf{0.0000e+0 (0.00e+0) =} & 0.0000e+0 (0.00e+0) = & 0.0000e+0 (0.00e+0) = & 0.0000e+0 (0.00e+0) = & 0.0000e+0 (0.00e+0) = & 0.0000e+0 (0.00e+0) = & 0.0000e+0 (0.00e+0) = & 0.0000e+0 (0.00e+0) = & 0.0000e+0 (0.00e+0) = & 0.0000e+0 (0.00e+0) \\
    RWCMOP21 & 80    & 2     & 6     & 20000 & 3.1699e-2 (1.09e-4) + & 3.1597e-2 (6.78e-5) - & 2.9323e-2 (2.41e-6) - & 3.1728e-2 (5.49e-5) + & 3.1529e-2 (8.83e-5) - & 3.1611e-2 (2.02e-5) - & 2.1361e-2 (2.45e-3) - & 3.0881e-2 (6.11e-5) - & \textbf{3.1761e-2 (2.80e-6) +} & 3.1640e-2 (7.55e-5) \\
    RWCMOP22 & 80    & 2     & 9     & 20000 & 0.0000e+0 (0.00e+0) - & 0.0000e+0 (0.00e+0) - & 0.0000e+0 (0.00e+0) - & 0.0000e+0 (0.00e+0) - & \textbf{8.5726e-1 (0.00e+0) =} & 8.0258e-1 (2.23e-1) = & 0.0000e+0 (0.00e+0) - & 5.7605e-1 (3.76e-1) = & 0.0000e+0 (0.00e+0) - & 8.2955e-1 (2.27e-1) \\
    RWCMOP23 & 80    & 2     & 6     & 20000 & 2.9213e-1 (1.52e-1) - & 0.0000e+0 (0.00e+0) - & 3.0561e-1 (1.18e-1) - & 4.0835e-1 (1.31e-1) - & 0.0000e+0 (0.00e+0) - & 8.9106e-1 (5.64e-2) - & 4.5356e-1 (2.65e-1) - & \textbf{9.9856e-1 (4.52e-16) =} & 2.5223e-1 (1.20e-1) - & 9.9856e-1 (4.52e-16) \\
    RWCMOP24 & 105   & 3     & 9     & 26250 & 0.0000e+0 (0.00e+0) - & \textbf{1.0061e+0 (0.00e+0)} & 0.0000e+0 (0.00e+0) - & 0.0000e+0 (0.00e+0) - & 0.0000e+0 (0.00e+0) - & 8.2141e-1 (2.98e-1) & 9.9996e-1 (0.00e+0) & 1.0000e+0 (0.00e+0) & 0.0000e+0 (0.00e+0) - & 0.0000e+0 (0.00e+0) - \\
    RWCMOP25 & 80    & 2     & 2     & 20000 & 2.4110e-1 (1.44e-5) + & 2.3976e-1 (3.02e-3) = & 2.3683e-1 (2.15e-4) - & 2.4117e-1 (1.38e-5) + & 2.4085e-1 (8.73e-5) = & 2.4075e-1 (7.40e-5) = & 1.9504e-1 (1.80e-2) - & 2.3523e-1 (5.56e-4) - & \textbf{2.4118e-1 (1.21e-5) +} & 2.4078e-1 (1.83e-4) \\
    \midrule
    \multicolumn{5}{c}{+/-/=}             & 14/9/2 & 6/14/5 & 2/20/3 & 12/9/4 & 5/14/6 & 6/14/5 & 1/21/3 & 3/18/4 & 16/5/4 &  \\
    \bottomrule
    \end{tabular}%
    \end{adjustbox}
\end{table*}%
\end{landscape}

\begin{figure*}[htbp]
	%\vspace{5pt}
	\centering
	\begin{subfigure}[b]{0.32\textwidth}
	\centering
			\includegraphics[width=1\textwidth]{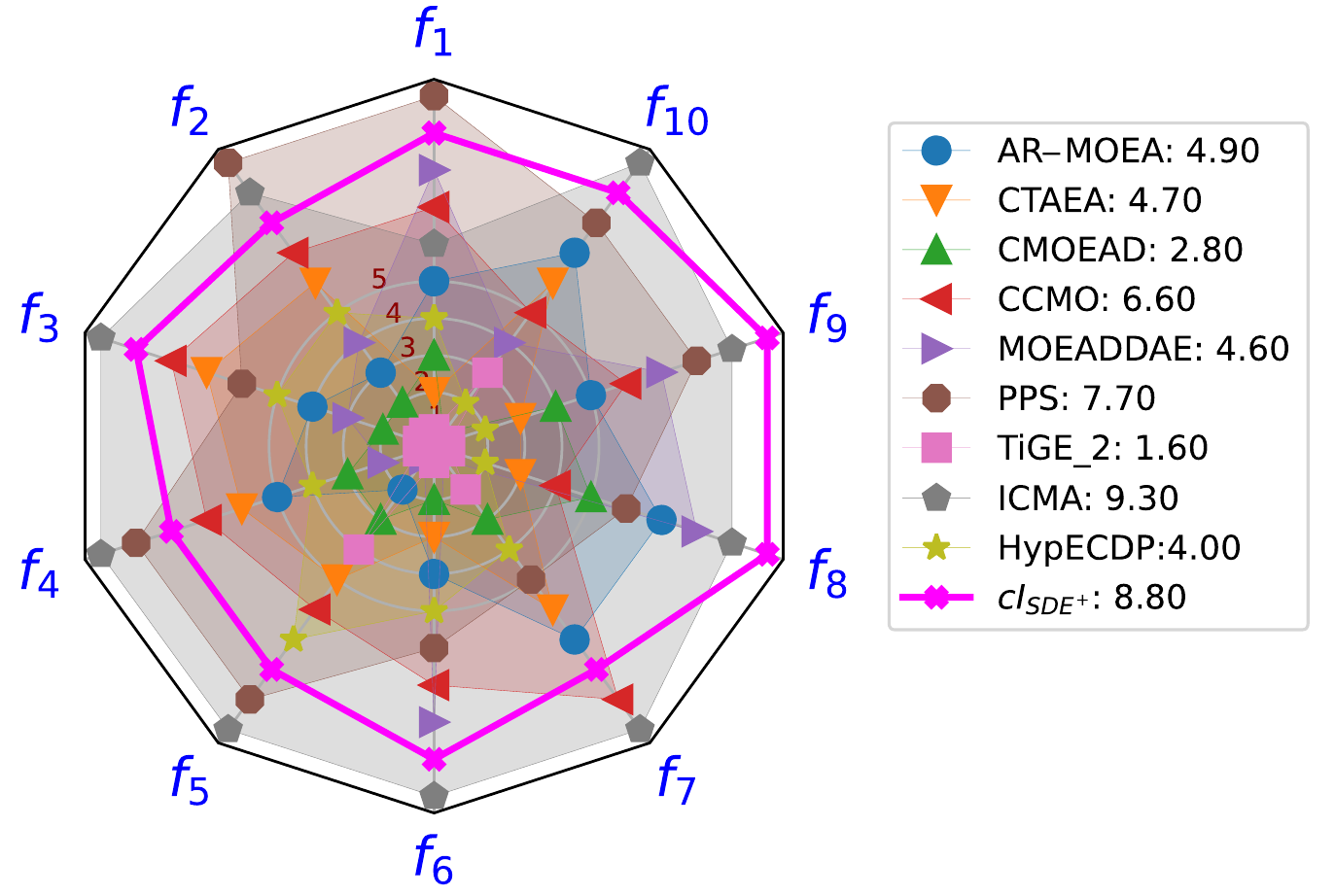}
			\caption{CF}	
			\label{fig:Ncovid}
		\end{subfigure}
		~
		\begin{subfigure}[b]{0.32\textwidth}
		\centering
			\includegraphics[width=1\textwidth]{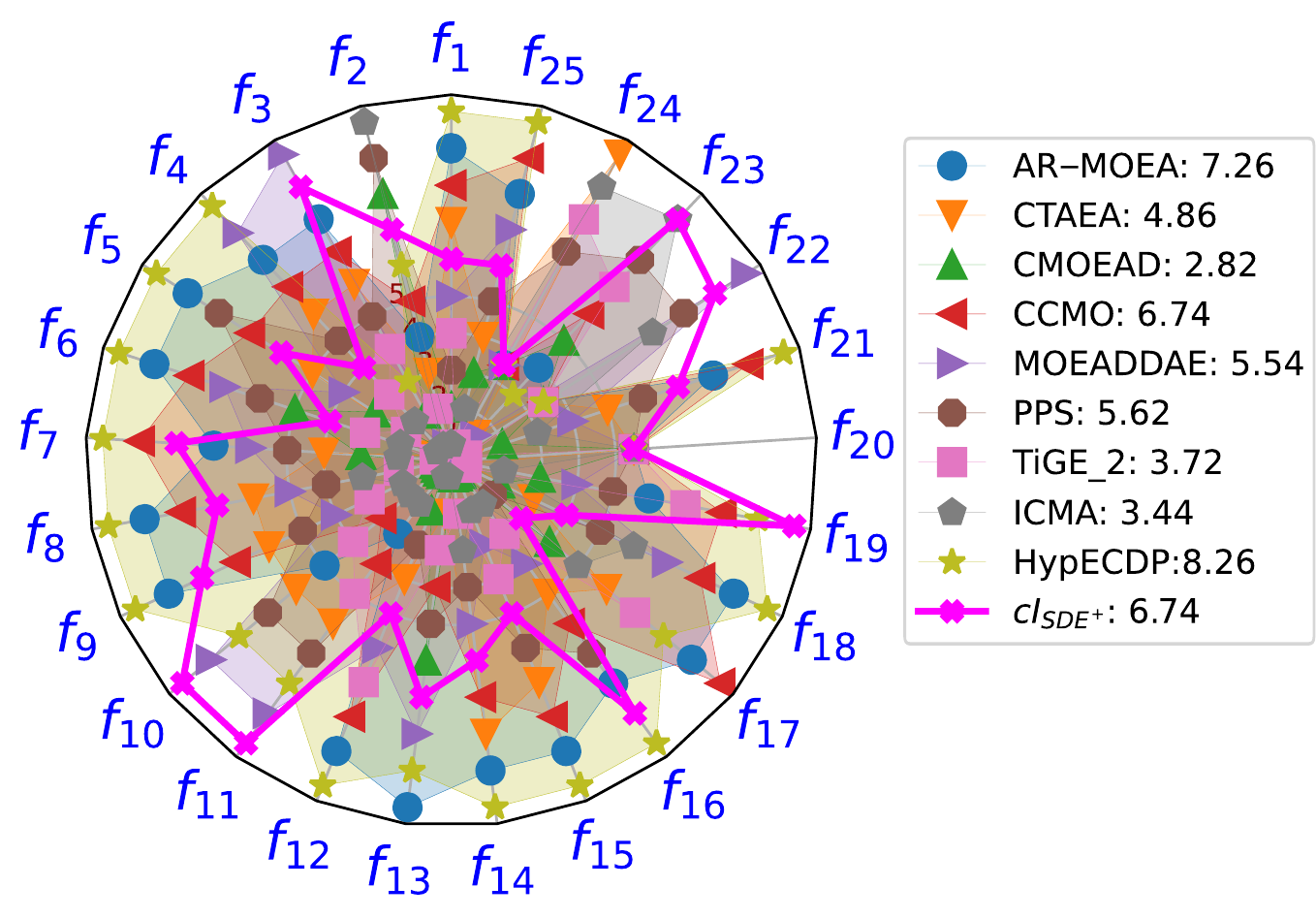}
			\caption{RWCMOP}
			\label{fig:Pcoovid}
	
		\end{subfigure}
		~
		\begin{subfigure}[b]{0.32\textwidth}
		\centering
			\includegraphics[width=1\textwidth]{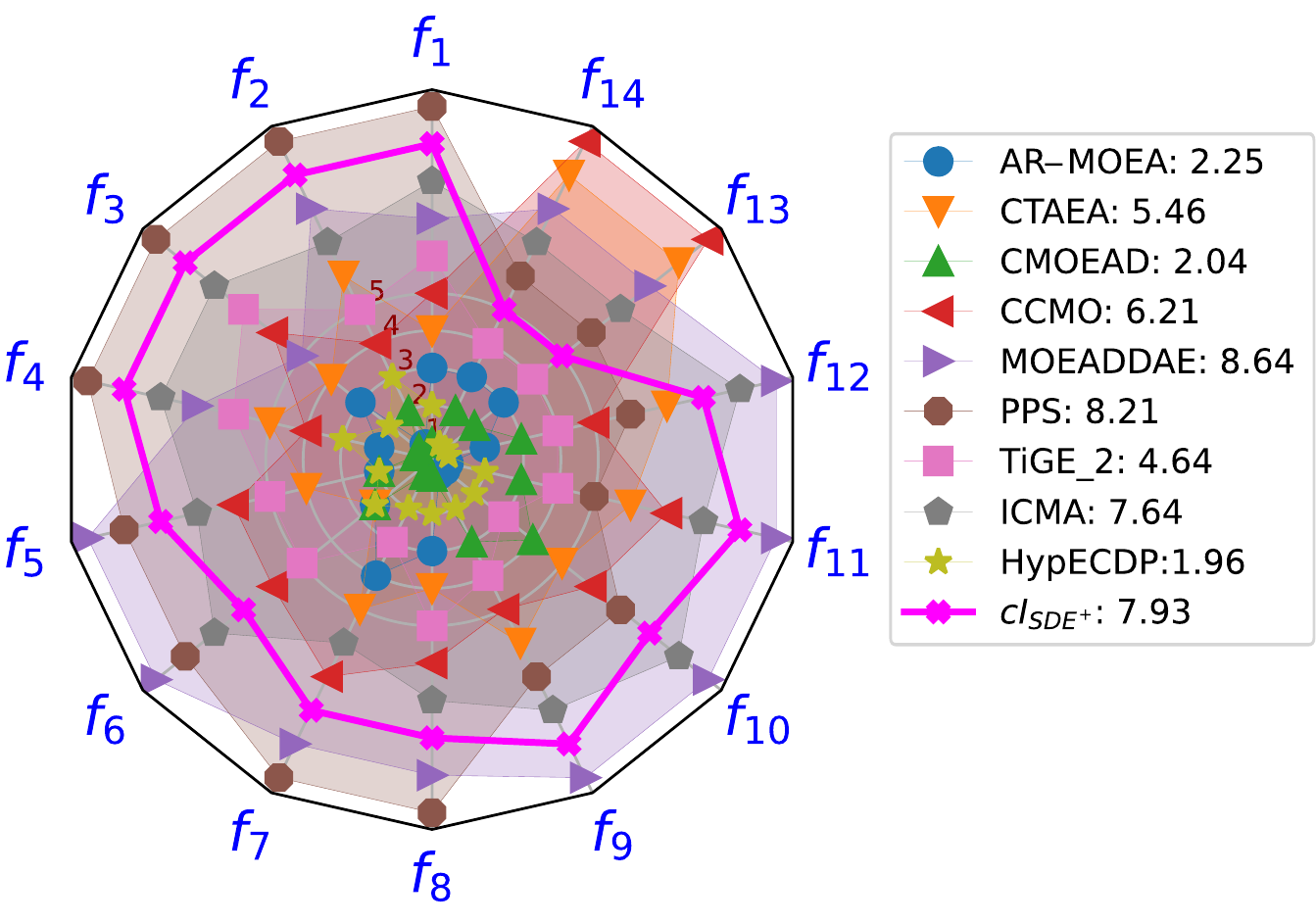}
			\caption{LIRCMOP}
			\label{fig:Normviz}
		\end{subfigure}
			~
		\begin{subfigure}[b]{0.32\textwidth}
		\centering
			\includegraphics[width=1\textwidth]{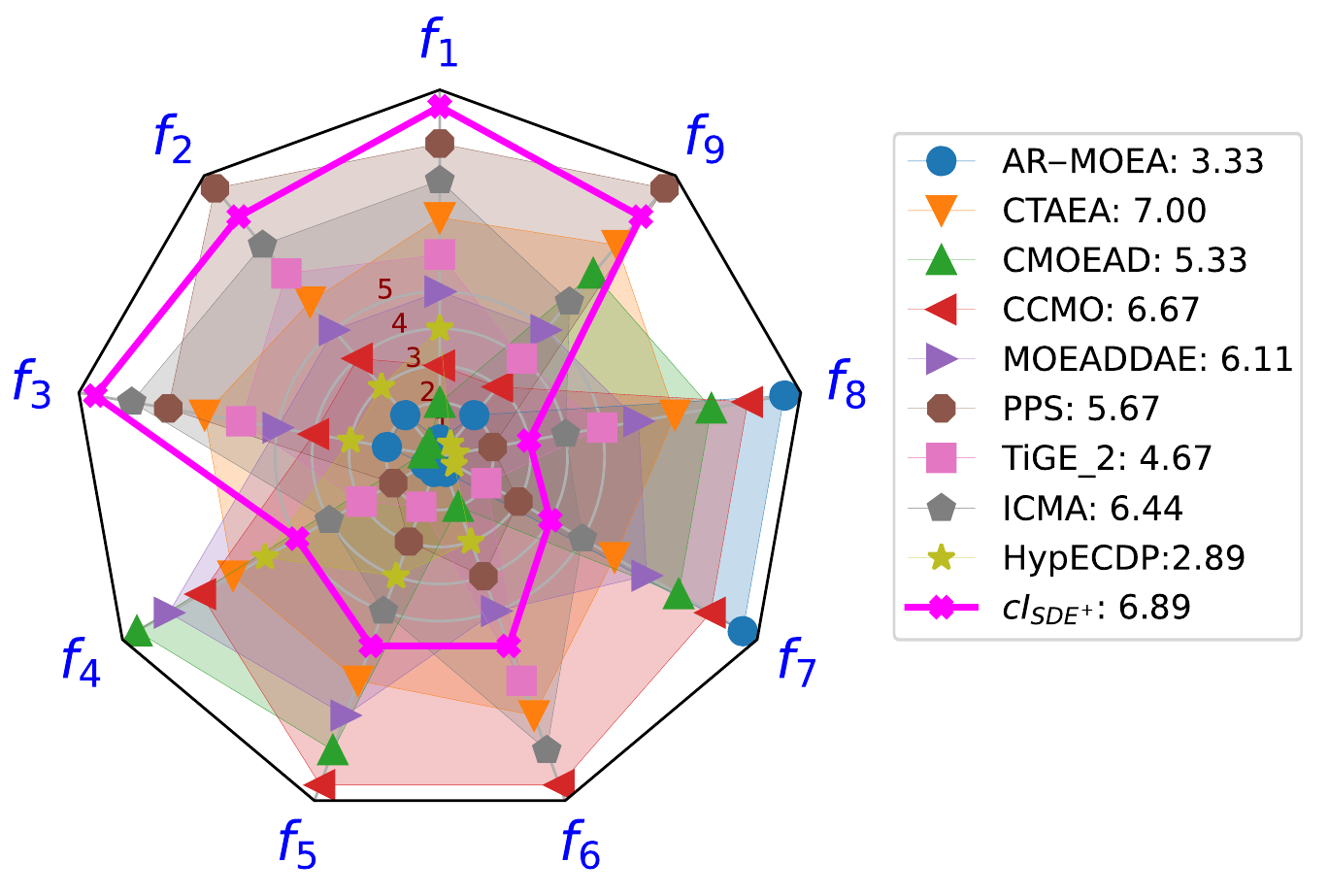}
			\caption{DASCMOP}
			\label{fig:Normviz}
		\end{subfigure}
					~
		\begin{subfigure}[b]{0.32\textwidth}
		\centering
			\includegraphics[width=1\textwidth]{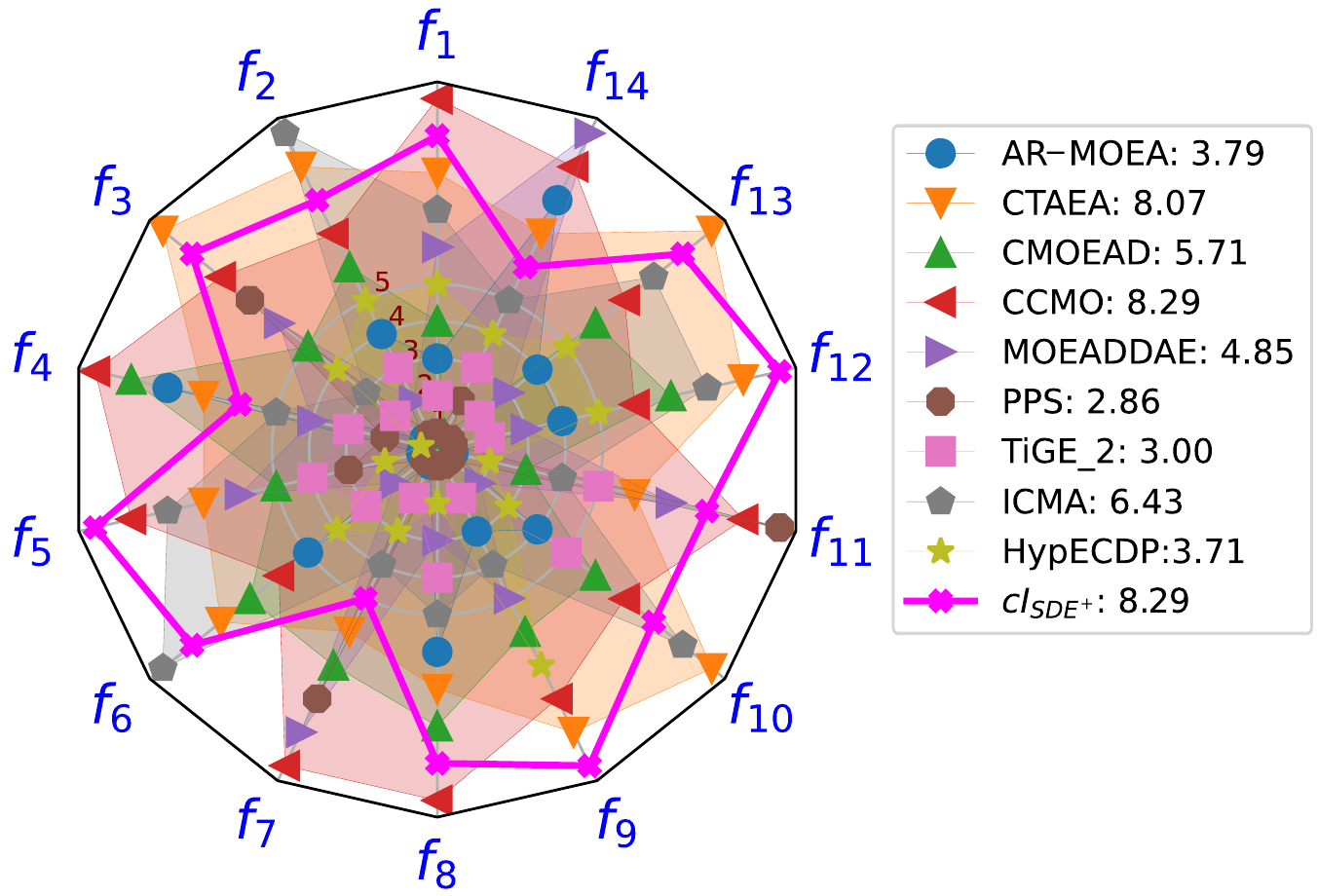}
			\caption{MW}
			\label{fig:Normviz}
		\end{subfigure}
			\begin{subfigure}[b]{0.32\textwidth}
		\centering
			\includegraphics[width=1\textwidth]{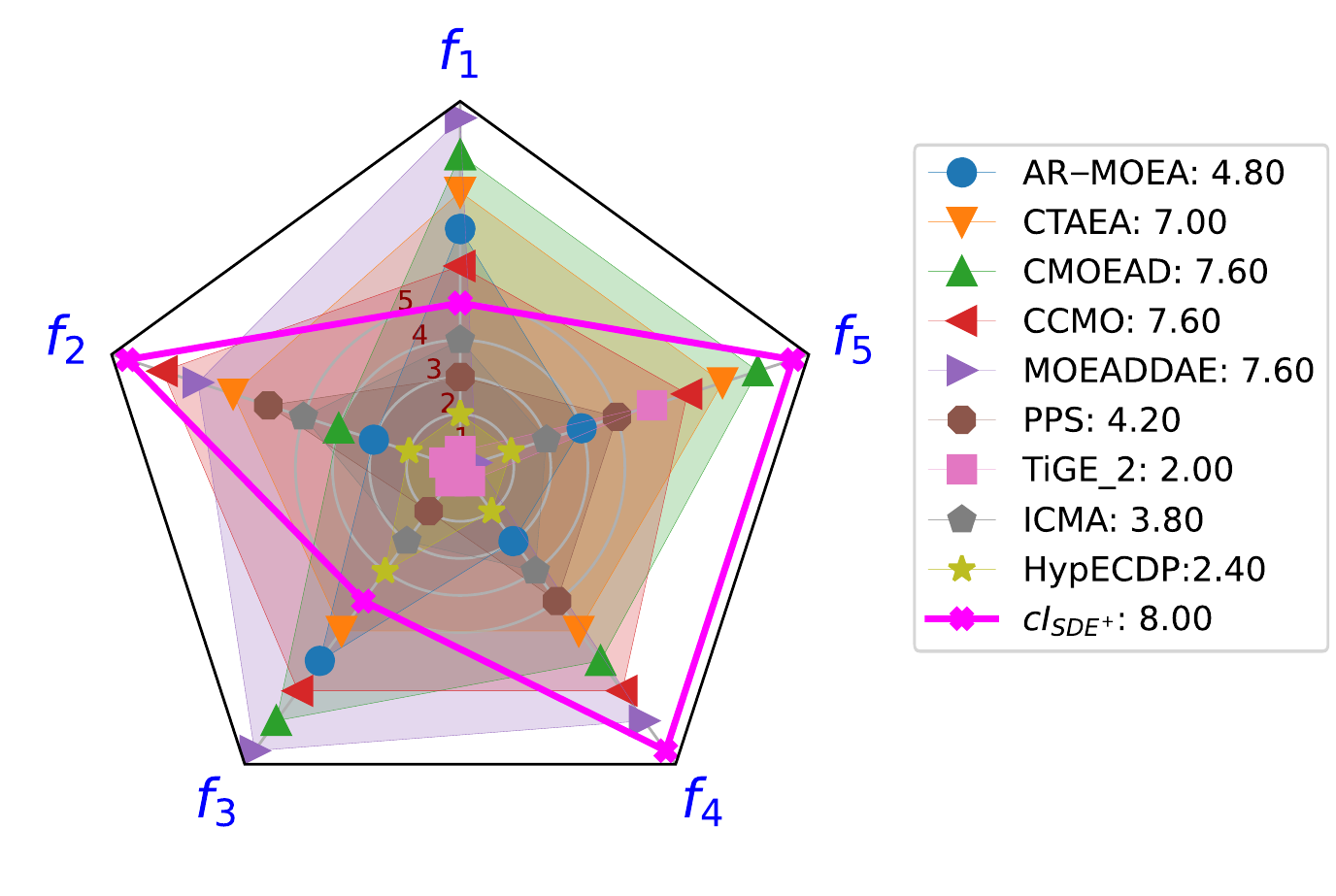}
			\caption{CDTLZ}
			\label{fig:Normviz}
		\end{subfigure}
	\caption{\textcolor{black}{Radar plots showing the average rank of CMOEAs on each test suite based on their HV values with Friedman Ranking in the legend}}
	\label{fig:radar}
 \end{figure*}

\color{black}
\textcolor{black}{The average ranks corresponding to each CMOEA on different test suites based on the mean HV values (radar plots) and Friedman ranking (legend) are presented in \textcolor{black}{Fig. \ref{fig:radar}}}. As CMOEAs are ranked in ascending order of HV values, CMOEA with largest HV is given a larger rank. Therefore, in the radar plots of Fig. \ref{fig:radar}, CMOEA covering larger volume is considered to be better. \textcolor{black}{During ranking, CMOEAs that have equal mean HV values, the average rank is considered. Table \ref{tab:F-rank} presents the Friedman ranking results on each test suite as well as the average Friedman ranks with scores. In Table \ref{tab:F-rank}, the values corresponding the best performance are highlighted.}
\textcolor{black}{All the statistical tests conducted in this work were performed with a p-value of 0.05.}

\begin{table}[h!]
  \centering
  \caption{Friedman ranking results of CMOEAs on each test suite as well as the average Friedman ranks with scores}
  \begin{adjustbox}{width=0.6\textwidth,center}
     \begin{tabular}{lcccccccc}
    \toprule
    \multirow{2}[2]{*}{Algorithms} & \multicolumn{6}{c}{Benchmark Test suites}             & \multicolumn{1}{c}{\multirow{2}[2]{*}{Average Rank}} & \multicolumn{1}{c}{\multirow{2}[2]{*}{Scores}} \\
 \cmidrule{2-7}  & CF    & RWCMOP & LIRCMOP & DASCMOP & MW    & CDTLZ &       &  \\
    \midrule
    AR-MOEA & 4.90   & 7.26  & 2.25  & 3.33  & 3.79  & 4.80   & 4.388 & 7 \\
    CTAEA & 4.7   & 4.86  & 5.46  & \textbf{7 .00}   & 8.07  & 7 .00    & 6.182 & 4 \\
    CMOEAD & 2.80   & 2.82  & 2.04  & 5.33  & 5.71  & 7.6 0  & 4.383 & 8 \\
    CCMO  & 6.60   & 6.74  & 6.21  & 6.67  & \textbf{8.29}  & 7.60   & 7.018 & 2 \\
    MOEADDAE & 4.60   & 5.54  & \textbf{8.64}  & 6.11  & 4.85  & 7.60   & 6.223 & 3 \\
    PPS   & 7.70   & 5.62  & 8.21  & 5.67  & 2.86  & 4.20   & 5.710 & 6 \\
    TiGE  & 1.6   & 3.72  & 4.64  & 4.67  & 3.00     & 2.00     & 3.272 & 10 \\
    ICMA  & \textbf{9.3}   & 3.44  & 7.64  & 6.44  & 6.43  & 3.8   & 6.175 & 5 \\
    HypECDP & 4.00     & \textbf{8.26}  & 1.96  & 2.89  & 3.71  & 2.40  & 3.870 & 9 \\
    $I^{c}_{SDE^+}$ & 8.80   & 6.74  & 7.93  & 6.89  & \textbf{8.29}  & \textbf{8.00}     & \textbf{7.775} & \textbf{1} \\
    \bottomrule
    \end{tabular}%
    \end{adjustbox}
  \label{tab:F-rank}%
\end{table}%

CMOEAD demonstrates degraded performance on all the test suites, except CDTLZ. This can be attributed to premature convergence due to over-emphasis on feasible solutions.

Generally, multi-stage based CMOEAs demonstrate superior performance on LIRCMOP as they focus on CPF and UPF, separately in different stages. However, MOEADDAE demonstrates competitive performance on CDTLZ whereas PPS fails. On the other hand, PPS demonstrates competitive performance on CF whereas MOEADDAE fails. The performance difference can be attributed to the characteristics of different stages involved and the switching mechanism employed. However, the performance of MOEADDAE and PPS degrades considerably on MW, with average performance on RWCMOP and DASCMOP.

The performance of 2 multi-population based CMOEAs follow similar pattern as in both the cases the goal of the 2 populations is same with only difference in the mechanisms employed. As one of the population is designed to find the UPF, multi-population approaches demonstrate better performance on DASCMOP that feature broken fronts where CPF $\subset$ UPF. In addition, their performance is competitive to the best performing CMOEA on MW. However, their performance is average on test suites such as CF, RWCMOP, LIRCMOP and CTDLZ. Among the 2 multi-population based CMOEAs, CCMO seems to have advantage over CTAEA due to roles assigned to different populations and how they interact.

Among the state-of-the-art indicator-based CMOEAs, TIGE\_2 fails on all the test instances due to its inability to balance the 3 employed indicators. HypECDP and \textcolor{black}{AR-MOEA} also fail in all test suites, except on RWCMOP, where they are ranked 1 and 2. On the other hand, ICMA demonstrates superior performance on CF and an average performance on LIRCMOP and DASCMOP. However, ICMA demonstrates degreaded performance on RWCMOP, MW and CDTLZ. The inconsistent and degraded performance of state-of-the-art indicator-based CMOEAs can be attributed to the inability of the indicators - single or fusion of multiple indicators, in enforcing the required diversity and convergence pressure during the evolution process.

\begin{figure*}[htbp]
	%\vspace{5pt}
	\centering
	\begin{subfigure}[b]{0.32\textwidth}
	\centering
			\includegraphics[width=0.95\textwidth]{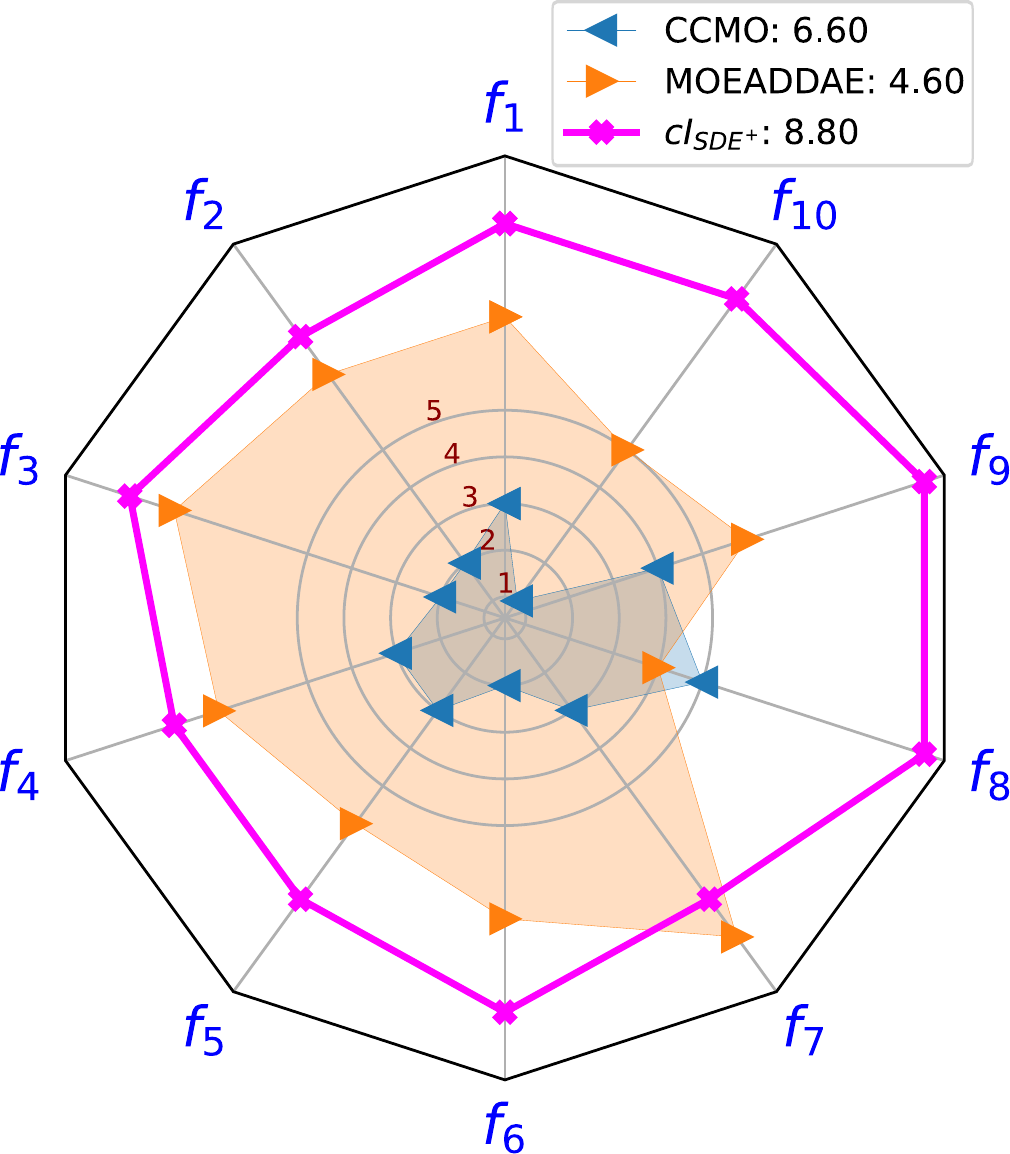}
			\caption{CF}	
			\label{fig:Ncovid}
		\end{subfigure}
		~
		\begin{subfigure}[b]{0.32\textwidth}
		\centering
			\includegraphics[width=0.95\textwidth]{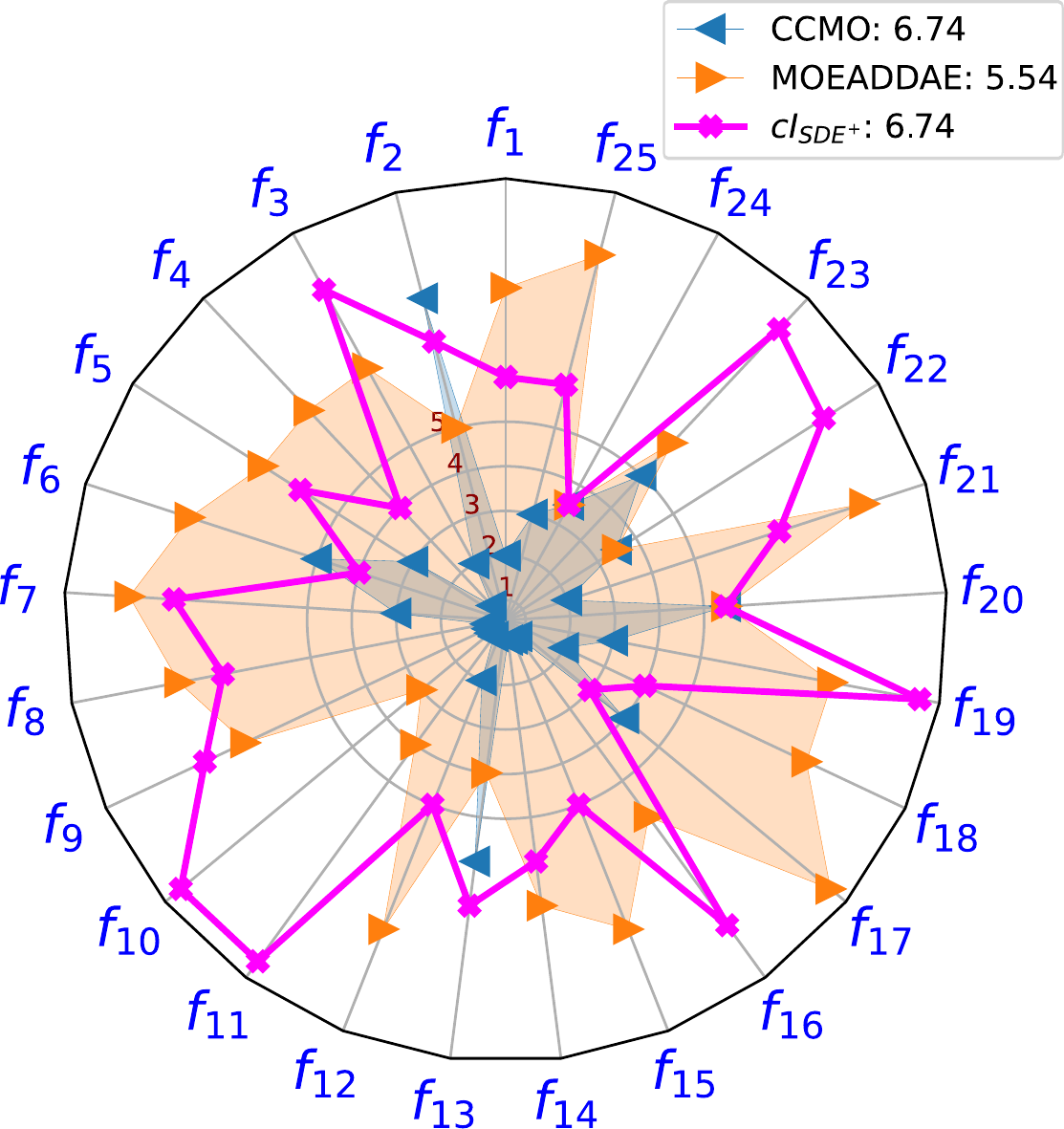}
			\caption{RWCMOP}
			\label{fig:Pcoovid}
	
		\end{subfigure}
		~
		\begin{subfigure}[b]{0.32\textwidth}
		\centering
			\includegraphics[width=0.95\textwidth]{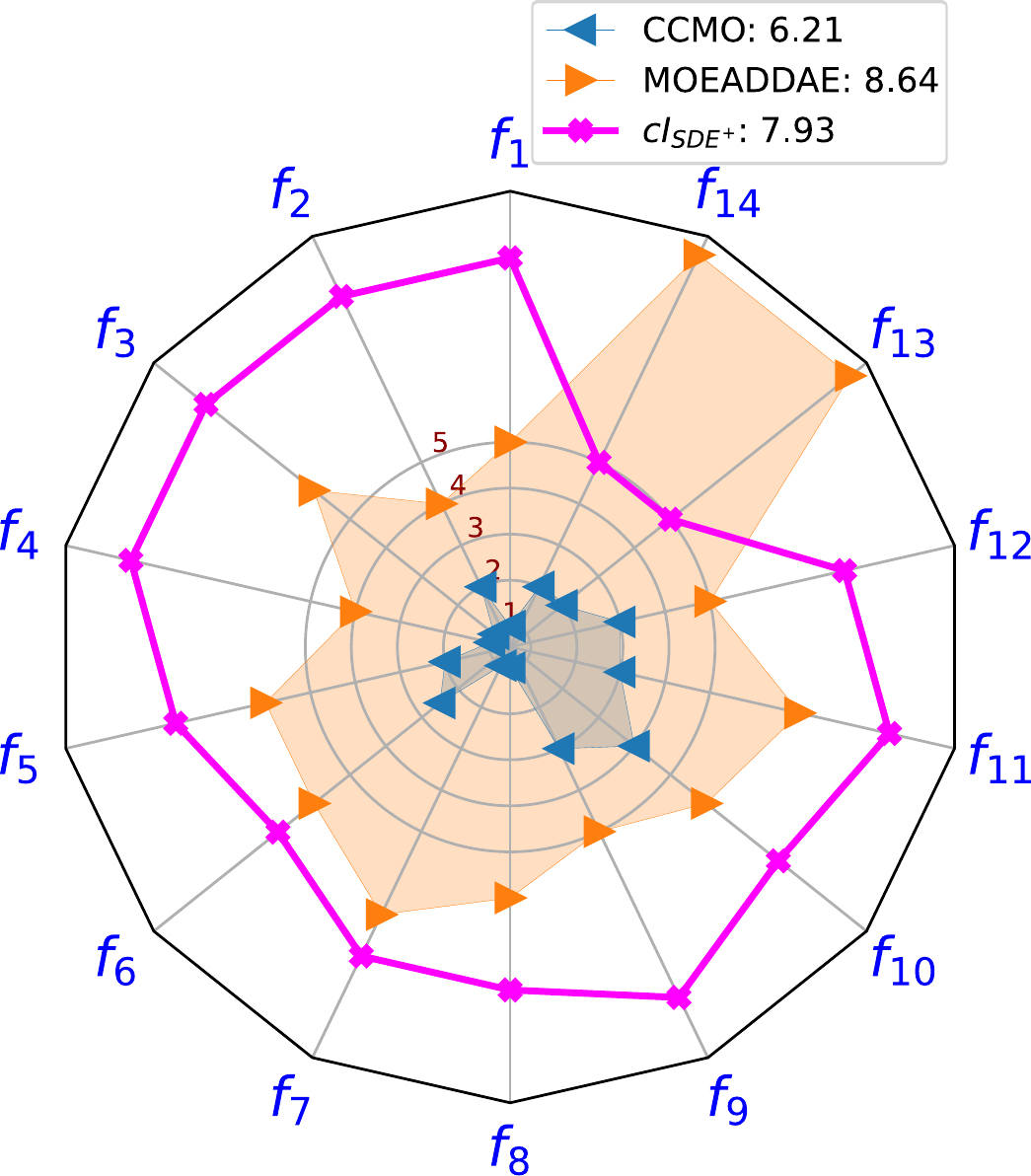}
			\caption{LIRCMOP}
			\label{fig:Normviz}
		\end{subfigure}
			~
		\begin{subfigure}[b]{0.32\textwidth}
		\centering
			\includegraphics[width=0.95\textwidth]{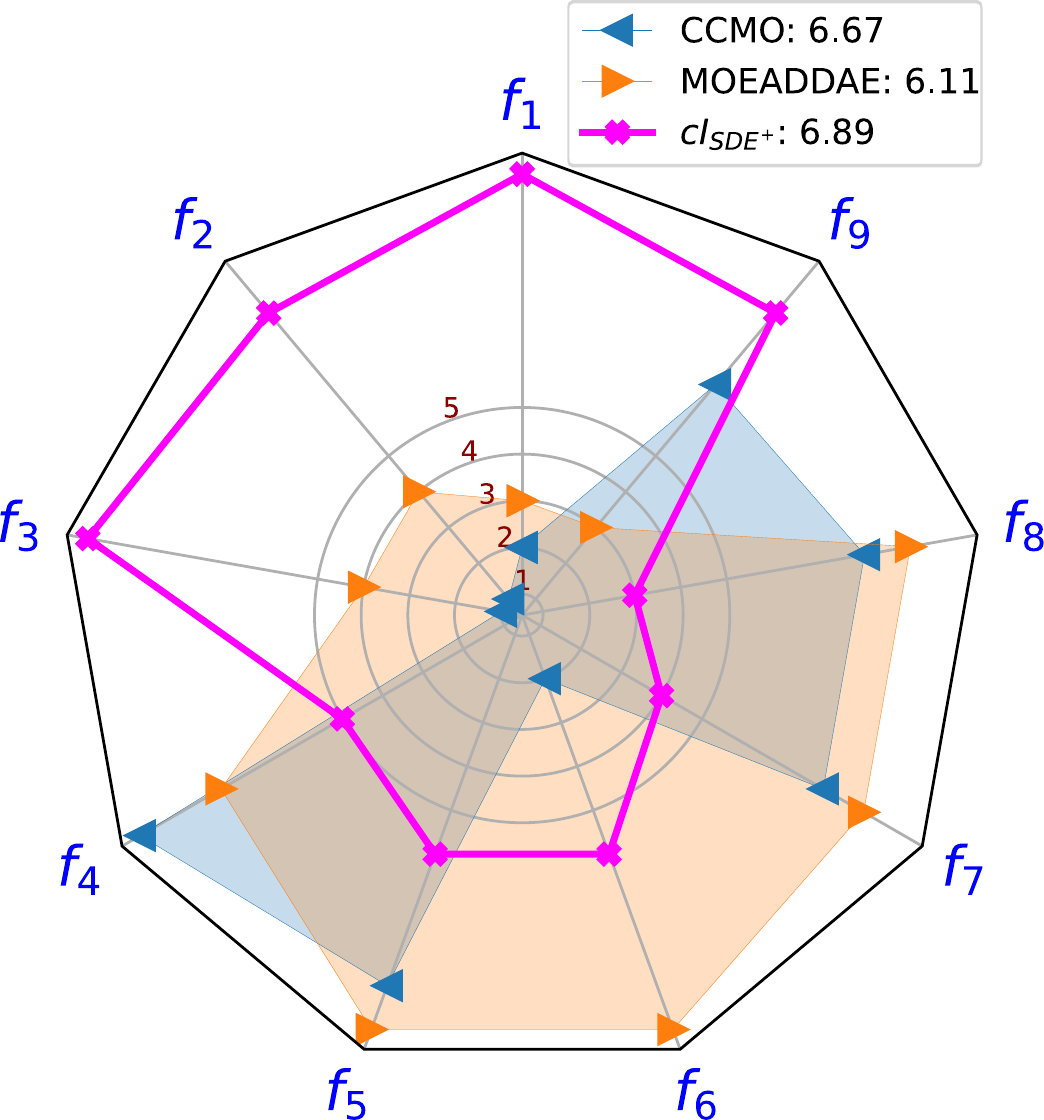}
			\caption{DASCMOP}
			\label{fig:Normviz}
		\end{subfigure}
					~
		\begin{subfigure}[b]{0.32\textwidth}
		\centering
			\includegraphics[width=0.95\textwidth]{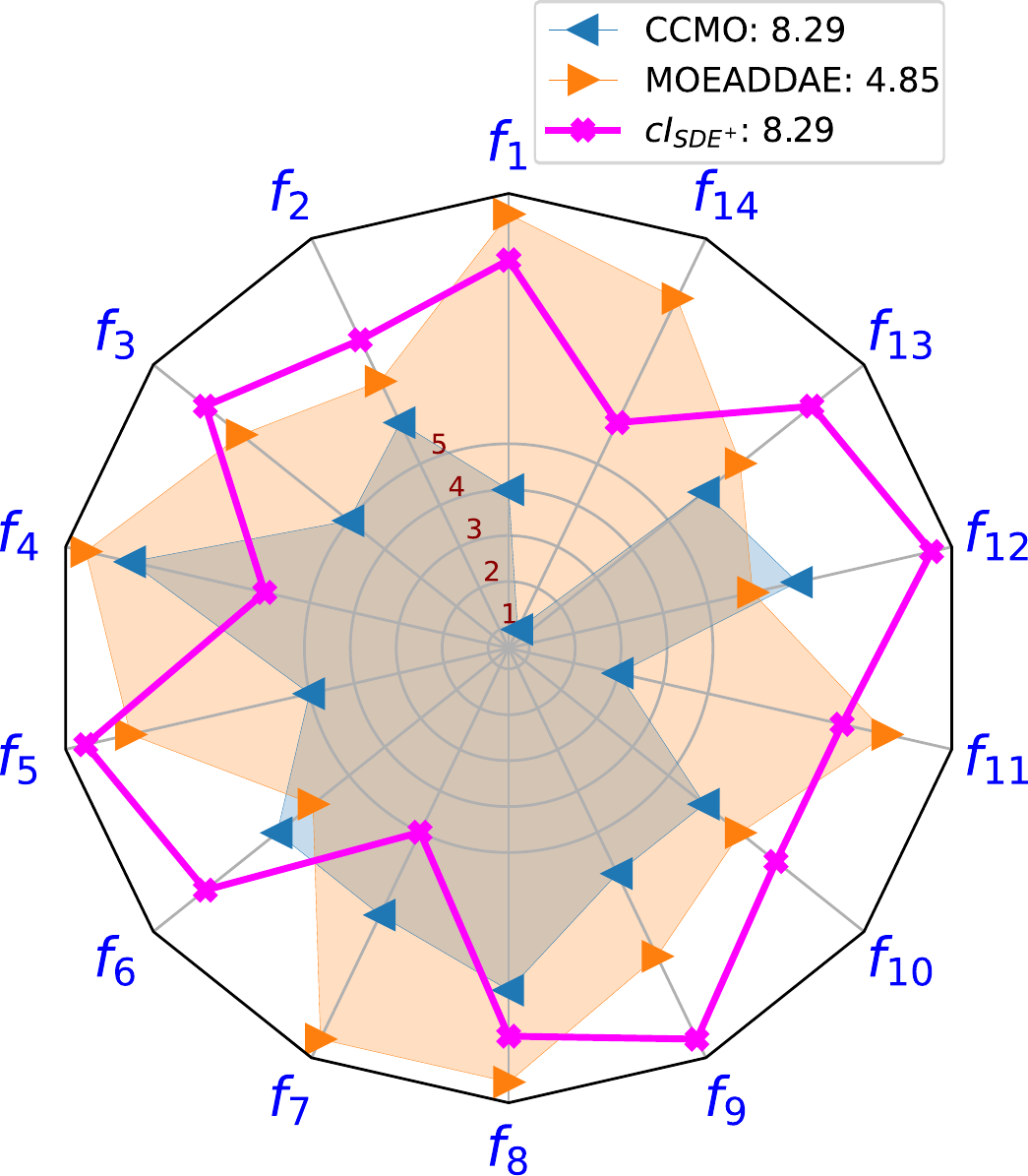}
			\caption{MW}
			\label{fig:Normviz}
		\end{subfigure}
			\begin{subfigure}[b]{0.32\textwidth}
		\centering
			\includegraphics[width=0.95\textwidth]{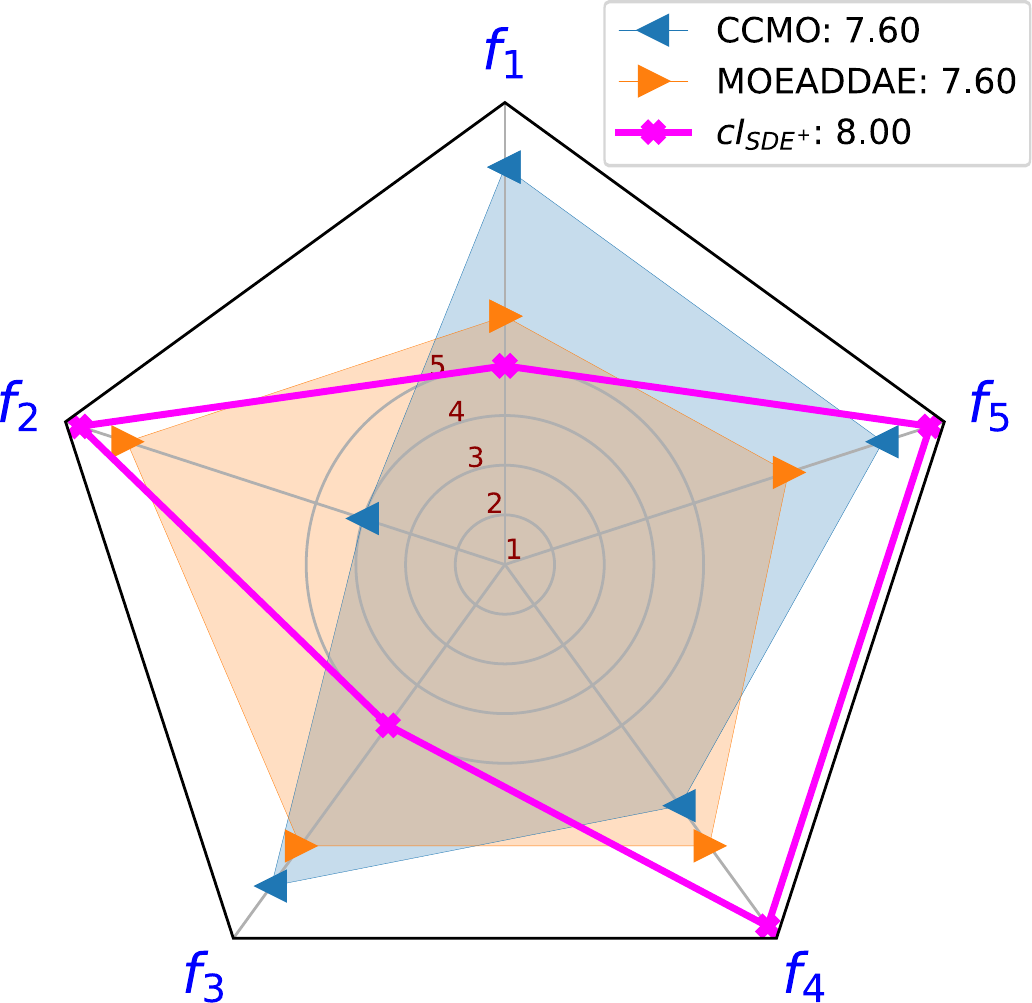}
			\caption{CDTLZ}
			\label{fig:Normviz}
		\end{subfigure}
	\caption{\textcolor{black}{Radar plots showing the average rank of top 3 ranked CMOEAs on each test suite based on their HV values with Friedman Ranking in the legend}}
	\label{fig:radar}
 \end{figure*}

\begin{figure}[h]
	%\vspace{5pt}
	\centering
	\begin{subfigure}[b]{0.8\textwidth}
	\centering
			\includegraphics[width=1\textwidth]{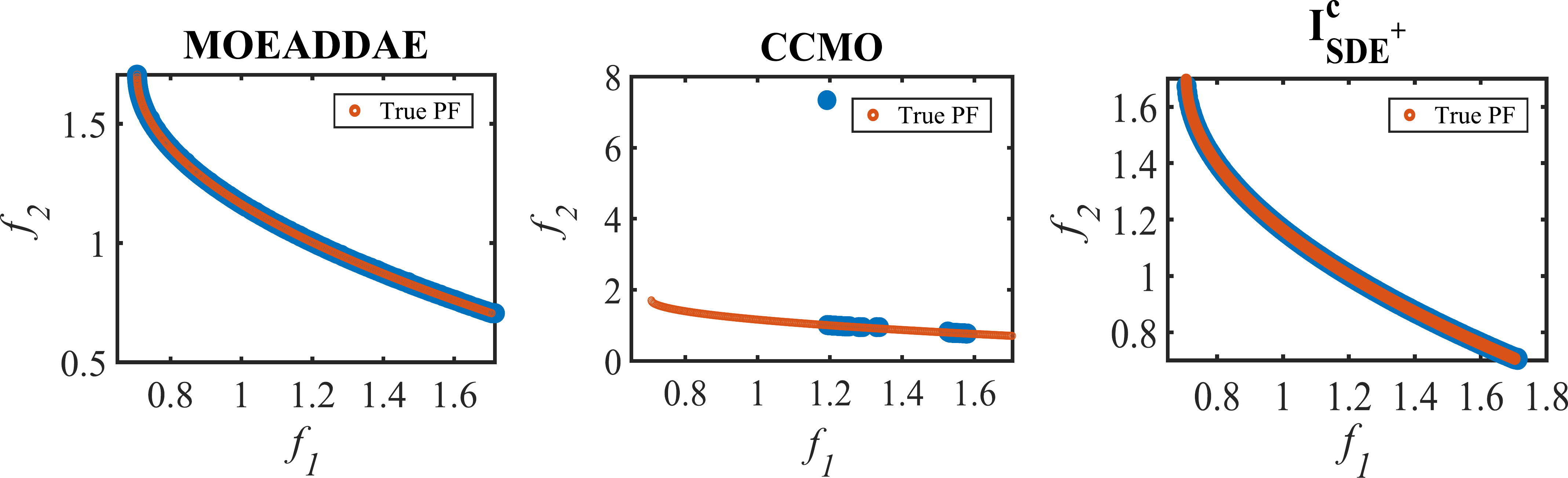}
			\caption{LIRCMOP 5}	
			\label{fig:Ncovid}
		\end{subfigure}
		~
		\begin{subfigure}[b]{0.8\textwidth}
		\centering
			\includegraphics[width=1\textwidth]{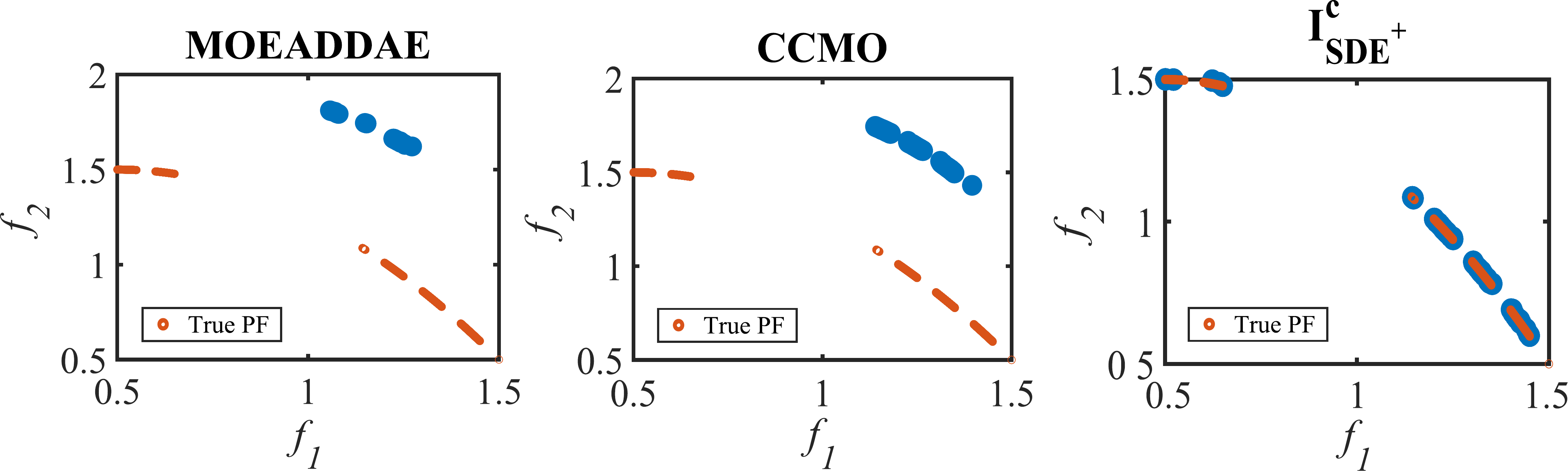}
			\caption{DASCMOP 1}
			\label{fig:Pcoovid}
		\end{subfigure}
		~
		\begin{subfigure}[b]{0.8\textwidth}
		\centering
			\includegraphics[width=1\textwidth]{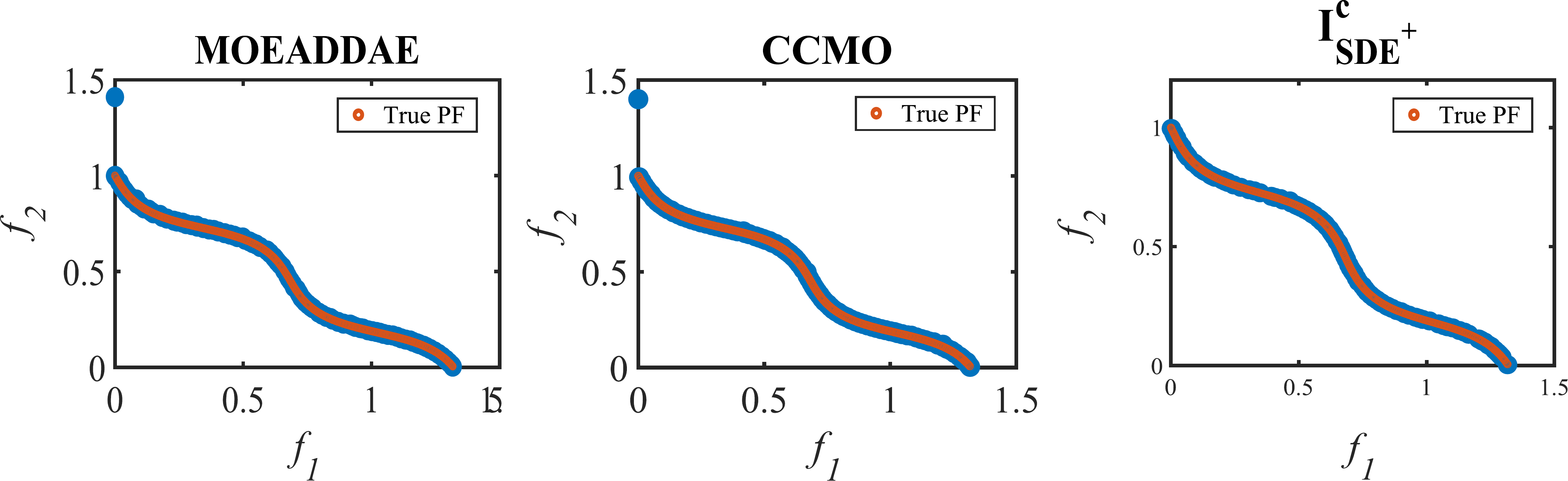}
			\caption{MW 12}
			\label{fig:Normviz}
		\end{subfigure}
	\caption{Performance comparison of Top 3 ranked CMOEAs in terms of final Pareto Fronts (PFs) on a) LIRCMOP5, b) DASCMOP1, and c) MW12}
	\label{fig:parallel_cord}
\end{figure}

From the results, it is evident that none of the state-of-the-art CMOEAs show consistent performance on different benchmark suites. Elevated performance on one is followed by degraded performance on the others. However, $I^{c}_{SDE^+}$ demonstrates consistent performance as it is always one of the best 3 performing CMOEAs on all the test suites. In other words, it ranked 2, 3, 3, 2, 1 and 1 on CF, RWCMOP, LIRCMOP, DASCMOP, MW and CDTLZ, respectively. \textcolor{black}{For qualitative analysis, Fig. \ref{fig:parallel_cord} presents the Pareto Fronts (PFs) corresponding to the final population obtained by Top 2 ranked CMOEAs. From Fig. \ref{fig:parallel_cord}, the consistent performance of $I^{c}_{SDE^+}$ in terms of convergence and spread compared to the CCMO \cite{ccmo} and MOEADDAE \cite{moeaddae} is evident.} The consistent performance of $I^{c}_{SDE^+}$ is due to its ability to efficiently handle the different discontinuities and explore all the feasible regions in the search space, due to the efficient fusion of the 3 components that complement each other. 
\subsection{Runtime Analysis}
In general, the computational complexity of evolutionary algorithms is expressed in terms of the average CPU time (t) in seconds for a single run taken by the algorithm. This is because EAs are stochastic and it becomes difficult to compute their asymptotic complexity.   In \citep{isde}, it was already proven experimentally that $I_{SDE^+}$ is computationally efficient compared to other SOTA MOEAS. Consequently, the proposed algorithm is motivated as a computationally efficient algorithm because the associated constraints handling method (superior of feasible) is simple and efficient. To evaluate the complexity of the proposed $I^{c}_{SDE^+}$, we presented the average CPU runtime and statistical comparison based on the Wilcoxon signed rank test of the top 3 performing algorithms featured in this work. The results are presented in Tables B15 to B20 in the appendix Section. From the results, it can be observed that the proposed algorithm is significantly computationally efficient compared with its top contenders across all the benchmark suites. 
\section{Conclusions and Future Directions}
A simple and efficient $I^{c}_{SDE^+}$ is proposed for CMOPs. The effectiveness of CMOEA with $I^{c}_{SDE^+}$, a single population-based framework, is demonstrated in comparison to the state-of-the-art CMOEAs. Formulated as a generalized fitness assignment method, $I^{c}_{SDE^+}$ can be extended for unconstrained multi-objective optimization by simply setting constraint violation to zero. Even though CMOEA with $I^{c}_{SDE^+}$ demonstrates significant performance, it fails to obtain corner solutions when solving complex CMOPs. This might be due to the use of SOB that concentrates the search around certain portions of the CPF, neglecting the corners. Instead of SOB, employing WSOB and adapting the weights during the evolution can help CMOEA cover the entire CPF.

\section*{Appendix}
\renewcommand\thefigure{\thesection\arabic{figure}}
\renewcommand\thetable{\thesection\arabic{table}}

\appendix 
\section{Comparison of the proposed $I^{c}_{SDE^+}$ with DE and GA Operators}
\label{sec:sample:appendix}
\textcolor{black}{In this Section, results of the comparison of the proposed $I^{c}_{SDE^+}$ algorithm using  DE and GA operators respectively are presented.}

\FloatBarrier
% Table generated by Excel2LaTeX from sheet 'HV'
\begin{table}[H]
  \centering
  \color{black}
  \caption{Comparison of the proposed $I^{c}_{SDE^+}$  with DE and GA Operators on CF test suite}
   \begin{adjustbox}{width=0.55\textwidth,center}
    \begin{tabular}{ccccccc}
    \toprule
    Problem & N     & M     & D     & FEs   &  $I^{c}_{SDE^+}$GA &  $I^{c}_{SDE^+}$DE \\
    \midrule
    CF1   & 100   & 2     & 10    & 300000 & 5.6203e-1 (3.06e-4) - & \textbf{5.6477e-1 (1.51e-4)} \\
    CF2   & 100   & 2     & 10    & 300000 & 6.3331e-1 (1.85e-2) - & \textbf{6.7000e-1 (6.64e-3)} \\
    CF3   & 100   & 2     & 10    & 300000 & 2.1018e-1 (3.42e-2) = & \textbf{2.1433e-1 (3.71e-2)} \\
    CF4   & 100   & 2     & 10    & 300000 & 4.2810e-1 (2.71e-2) - & \textbf{4.6880e-1 (1.63e-2)} \\
    CF5   & 100   & 2     & 10    & 300000 & 2.7374e-1 (6.08e-2) - & \textbf{3.1160e-1 (5.34e-2)} \\
    CF6   & 100   & 2     & 10    & 300000 & 6.5870e-1 (1.15e-2) - & \textbf{6.8133e-1 (1.22e-2)} \\
    CF7   & 100   & 2     & 10    & 300000 & 4.2908e-1 (8.76e-2) - & \textbf{4.7125e-1 (1.11e-1)} \\
    CF8   & 150   & 3     & 10    & 300000 & 4.4663e-1 (2.00e-2) - & \textbf{5.0122e-1 (1.97e-2)} \\
    CF9   & 150   & 3     & 10    & 300000 & 4.7492e-1 (1.54e-2) - & \textbf{5.3216e-1 (1.09e-2)} \\
    CF10  & 150   & 3     & 10    & 300000 & 2.7072e-1 (1.00e-1) = & \textbf{3.4766e-1 (9.26e-2)} \\
    \midrule
    \multicolumn{5}{c}{+/-/=}             & 0/8/2 &  \\
    \bottomrule
    \end{tabular}%
    \end{adjustbox}
  \label{tab:addlabel}%
\end{table}%

% Table generated by Excel2LaTeX from sheet 'HV'
\begin{table}[H]
  \centering
  \color{black}
  \caption{Comparison of the proposed $I^{c}_{SDE^+}$  with DE and GA Operators on CDTLZ test suite}
   \begin{adjustbox}{width=0.55\textwidth,center}
    \begin{tabular}{cccccc}
    \toprule
    Problem & N     & M     & D     &  $I^{c}_{SDE^+}$DE &  $I^{c}_{SDE^+}$GA \\
    \midrule
    C1\_DTLZ1 & 92    & 3     & 7     & 8.2319e-1 (1.41e-2) = & 8.2892e-1 (4.48e-3) \\
    C1\_DTLZ3 & 92    & 3     & 12    & 2.4815e-1 (2.62e-1) - & \textbf{5.5974e-1 (1.29e-3)} \\
    C2\_DTLZ2 & 92    & 3     & 12    & 4.9236e-1 (3.45e-3) - & 5.0214e-1 (3.02e-3) \\
    C3\_DTLZ1 & 92    & 3     & 7     & 3.1422e-1 (7.10e-3) = & \textbf{3.1650e-1 (9.61e-3)} \\
    C3\_DTLZ4 & 92    & 3     & 12    & 7.9256e-1 (5.61e-4) - & \textbf{7.9352e-1 (4.45e-4)} \\
    \midrule
    \multicolumn{4}{c}{+/-/=}     & 0/3/2 &  \\
    \bottomrule
    \end{tabular}%
    \end{adjustbox}
  \label{tab:addlabel}%
\end{table}%

% Table generated by Excel2LaTeX from sheet 'HV'
\begin{table}[H]
  \centering
  \color{black}
  \caption{Comparison of the proposed $I^{c}_{SDE^+}$  with DE and GA operators on DASCMOP test suite}
   \begin{adjustbox}{width=0.55\textwidth,center}
    \begin{tabular}{ccccccc}
    \toprule
    Problem & N     & M     & D     & FEs   &  $I^{c}_{SDE^+}$GA &  $I^{c}_{SDE^+}$DE \\
    \midrule
    DASCMOP1 & 300   & 2     & 30    & 300000 & 3.2242e-2 (2.60e-2) - & \textbf{2.0013e-1 (2.41e-3)} \\
    DASCMOP2 & 300   & 2     & 30    & 300000 & 2.7390e-1 (5.91e-3) - & \textbf{3.4872e-1 (7.39e-4)} \\
    DASCMOP3 & 300   & 2     & 30    & 300000 & 2.5296e-1 (8.94e-3) - & \textbf{3.1180e-1 (1.59e-4)} \\
    DASCMOP4 & 300   & 2     & 30    & 300000 & \textbf{2.0065e-1 (2.91e-3) +} & 1.9713e-1 (1.39e-2) \\
    DASCMOP5 & 300   & 2     & 30    & 300000 & \textbf{3.4926e-1 (4.37e-4) +} & 3.4767e-1 (1.17e-3) \\
    DASCMOP6 & 300   & 2     & 30    & 300000 & \textbf{3.1197e-1 (2.10e-4) +} & 2.9833e-1 (2.88e-2) \\
    DASCMOP7 & 300   & 3     & 30    & 300000 & \textbf{2.8315e-1 (1.43e-3) +} & 2.7286e-1 (1.20e-2) \\
    DASCMOP8 & 300   & 3     & 30    & 300000 & \textbf{1.9574e-1 (4.13e-3) +} & 1.9236e-1 (4.78e-3) \\
    DASCMOP9 & 300   & 3     & 30    & 300000 & 1.5935e-1 (1.33e-2) - & \textbf{1.9556e-1 (4.76e-3)} \\
    \midrule
    \multicolumn{5}{c}{+/-/=}             & 5/4/0 &  \\
    \bottomrule
    \end{tabular}%
    \end{adjustbox}
  \label{tab:addlabel}%
\end{table}%

% Table generated by Excel2LaTeX from sheet 'HV'
\begin{table}[H]
  \centering
  \color{black}
  \caption{Comparison of the proposed $I^{c}_{SDE^+}$  with DE and GA Operators on LIRCMOP test suite}
   \begin{adjustbox}{width=0.55\textwidth,center}
    \begin{tabular}{ccccccc}
    \toprule
    Problem & N     & M     & D     & FEs   &  $I^{c}_{SDE^+}$GA &  $I^{c}_{SDE^+}$DE \\
    \midrule
    LIRCMOP1 & 300   & 2     & 30    & 300000 & 2.1590e-1 (4.09e-3) - & \textbf{2.2942e-1 (1.47e-3)} \\
    LIRCMOP2 & 300   & 2     & 30    & 300000 & 3.3442e-1 (3.19e-3) - & \textbf{3.5243e-1 (1.24e-3)} \\
    LIRCMOP3 & 300   & 2     & 30    & 300000 & 1.8837e-1 (4.12e-3) - & \textbf{1.9667e-1 (4.00e-3)} \\
    LIRCMOP4 & 300   & 2     & 30    & 300000 & 2.8590e-1 (6.00e-3) - & \textbf{3.0401e-1 (4.38e-3)} \\
    LIRCMOP5 & 300   & 2     & 30    & 300000 & 1.1208e-1 (6.45e-2) - & \textbf{2.9142e-1 (6.69e-4)} \\
    LIRCMOP6 & 300   & 2     & 30    & 300000 & 7.9137e-2 (4.13e-2) - & \textbf{1.8063e-1 (3.18e-2)} \\
    LIRCMOP7 & 300   & 2     & 30    & 300000 & 2.4384e-1 (4.64e-3) - & \textbf{2.9070e-1 (1.37e-3)} \\
    LIRCMOP8 & 300   & 2     & 30    & 300000 & 2.2745e-1 (4.42e-3) - & \textbf{2.8733e-1 (1.43e-2)} \\
    LIRCMOP9 & 300   & 2     & 30    & 300000 & 3.6030e-1 (7.67e-2) - & \textbf{5.5537e-1 (5.39e-3)} \\
    LIRCMOP10 & 300   & 2     & 30    & 300000 & 5.1292e-1 (1.12e-1) - & \textbf{7.0165e-1 (2.72e-3)} \\
    LIRCMOP11 & 300   & 2     & 30    & 300000 & 6.0289e-1 (1.18e-1) - & \textbf{6.9307e-1 (1.11e-3)} \\
    LIRCMOP12 & 300   & 2     & 30    & 300000 & 4.7424e-1 (5.88e-2) - & \textbf{6.0485e-1 (2.82e-2)} \\
    LIRCMOP13 & 300   & 3     & 30    & 300000 & \textbf{5.3215e-1 (9.37e-3) +} & 5.2426e-1 (9.31e-3) \\
    LIRCMOP14 & 300   & 3     & 30    & 300000 & 5.2867e-1 (1.06e-2) = & \textbf{5.2999e-1 (5.99e-3)} \\
    \midrule
    \multicolumn{5}{c}{+/-/=}             & 1/12/1 &  \\
    \bottomrule
    \end{tabular}%
    \end{adjustbox}
  \label{tab:addlabel}%
\end{table}%

% Table generated by Excel2LaTeX from sheet 'HV'
\begin{table}[H]
  \centering
  \color{black}
  \caption{Comparison of the proposed $I^{c}_{SDE^+}$  with DE and GA Operators on MW test suite}
   \begin{adjustbox}{width=0.55\textwidth,center}
    \begin{tabular}{ccccccc}
    \toprule
    Problem & N     & M     & D     & FEs   &  $I^{c}_{SDE^+}$DE &  $I^{c}_{SDE^+}$GA \\
    \midrule
    MW1   & 100   & 2     & 15    & 60000 & 4.7160e-1 (3.74e-2) - & \textbf{4.8910e-1 (2.73e-4)} \\
    MW2   & 100   & 2     & 15    & 60000 & 4.0551e-1 (1.07e-1) - & \textbf{5.5905e-1 (1.23e-2)} \\
    MW3   & 100   & 2     & 15    & 60000 & \textbf{5.4425e-1 (6.16e-4) +} & 5.4390e-1 (5.96e-4) \\
    MW4   & 100   & 3     & 15    & 60000 & 8.2653e-1 (9.79e-3) - & \textbf{8.3803e-1 (1.38e-3)} \\
    MW5   & 100   & 2     & 15    & 60000 & 2.6426e-1 (9.98e-2) - & \textbf{3.2302e-1 (5.48e-4)} \\
    MW6   & 100   & 2     & 15    & 60000 & 6.9554e-2 (5.88e-2) - & \textbf{3.1217e-1 (1.37e-2)} \\
    MW7   & 100   & 2     & 15    & 60000 & \textbf{4.0747e-1 (1.02e-3) +} & 4.0655e-1 (1.42e-3) \\
    MW8   & 100   & 3     & 15    & 60000 & 3.1787e-1 (1.09e-1) - & \textbf{5.3305e-1 (1.08e-2)} \\
    MW9   & 100   & 2     & 15    & 60000 & 1.2974e-1 (1.79e-1) - & \textbf{3.9431e-1 (2.89e-3)} \\
    MW10  & 100   & 2     & 15    & 60000 & 2.0024e-1 (1.10e-1) - & \textbf{4.1623e-1 (1.90e-2)} \\
    MW11  & 100   & 2     & 15    & 60000 & 4.4439e-1 (5.86e-4) = & \textbf{4.4448e-1 (5.79e-4)} \\
    MW12  & 100   & 2     & 15    & 60000 & 1.4228e-1 (2.06e-1) - & \textbf{6.0397e-1 (3.97e-4)} \\
    MW13  & 100   & 2     & 15    & 60000 & 3.0556e-1 (9.60e-2) - & \textbf{4.5051e-1 (1.24e-2)} \\
    MW14  & 100   & 3     & 15    & 60000 & 4.5576e-1 (1.45e-2) - & \textbf{4.6430e-1 (6.80e-3)} \\
    \midrule
    \multicolumn{5}{c}{+/-/=}             & 2/11/1 &  \\
    \bottomrule
    \end{tabular}%
    \end{adjustbox}
  \label{tab:addlabel}%
\end{table}%

% Table generated by Excel2LaTeX from sheet 'HV'
\begin{table}[H]
  \centering
  \color{black}
  \caption{Comparison of the proposed $I^{c}_{SDE^+}$  with DE and GA Operators on RWCMOP test suite}
     \begin{adjustbox}{width=0.55\textwidth,center}
    \begin{tabular}{ccccccc}
    \toprule
    Problem & N     & M     & D     & FEs   &  $I^{c}_{SDE^+}$GA &  $I^{c}_{SDE^+}$DE \\
    \midrule
    RWMOP1 & 80    & 2     & 4     & 20000 & 5.9793e-1 (5.27e-3) - & \textbf{5.9815e-1 (1.61e-3)} \\
    RWMOP2 & 80    & 2     & 5     & 20000 & \textbf{3.5911e-1 (2.23e-2) +} & 2.7352e-1 (7.01e-2) \\
    RWMOP3 & 80    & 2     & 3     & 20000 & \textbf{9.0039e-1 (3.64e-4) +} & 9.0001e-1 (3.79e-4) \\
    RWMOP4 & 80    & 2     & 4     & 20000 & \textbf{8.4608e-1 (8.99e-3) +} & 8.3719e-1 (8.55e-3) \\
    RWMOP5 & 80    & 2     & 4     & 20000 & 4.2798e-1 (1.83e-3) - & \textbf{4.3154e-1 (9.27e-4)} \\
    RWMOP6 & 80    & 2     & 7     & 20000 & \textbf{2.7600e-1 (2.13e-4) +} & 2.7449e-1 (3.52e-4) \\
    RWMOP7 & 80    & 2     & 4     & 20000 & 4.8344e-1 (7.53e-4) - & \textbf{4.8403e-1 (5.21e-4)} \\
    RWMOP8 & 105   & 3     & 7     & 26250 & 2.5233e-2 (7.96e-4) - & \textbf{2.5819e-2 (5.54e-4)} \\
    RWMOP9 & 80    & 2     & 4     & 20000 & 4.0763e-1 (1.09e-3) - & \textbf{4.0916e-1 (2.33e-4)} \\
    RWMOP10 & 80    & 2     & 2     & 20000 & 8.4720e-1 (1.90e-4) - & \textbf{8.4729e-1 (1.14e-4)} \\
    RWMOP11 & 212   & 5     & 3     & 53000 & 1.0192e-1 (3.49e-4) - & \textbf{1.0213e-1 (2.36e-4)} \\
    RWMOP12 & 80    & 2     & 4     & 20000 & 5.4025e-1 (6.62e-3) - & \textbf{5.4663e-1 (6.38e-3)} \\
    RWMOP13 & 105   & 3     & 7     & 26250 & \textbf{8.9992e-2 (3.04e-4) +} & 8.9986e-2 (1.19e-4) \\
    RWMOP14 & 80    & 2     & 5     & 20000 & 6.1034e-1 (2.23e-3) - & \textbf{6.1238e-1 (2.19e-3)} \\
    RWMOP15 & 80    & 2     & 3     & 20000 & 4.8595e-1 (4.30e-2) - & \textbf{5.2379e-1 (1.47e-2)} \\
    RWMOP16 & 80    & 2     & 2     & 20000 & 7.6243e-1 (6.67e-4) - & \textbf{7.6274e-1 (5.27e-4)} \\
    RWMOP17 & 105   & 3     & 6     & 26250 & \textbf{2.5302e-1 (1.72e-2) +} & 2.3174e-1 (3.62e-2) \\
    RWMOP18 & 80    & 2     & 3     & 20000 & 4.0345e-2 (8.23e-5) = & \textbf{4.0365e-2 (7.61e-5)} \\
    RWMOP19 & 105   & 3     & 10    & 26250 & 3.3527e-1 (1.02e-2) - & \textbf{3.4772e-1 (4.49e-3)} \\
    RWMOP20 & 80    & 2     & 4     & 20000 & 0.0000e+0 (0.00e+0) = & 0.0000e+0 (0.00e+0) \\
    RWMOP21 & 80    & 2     & 6     & 20000 & 3.0903e-2 (5.96e-4) - & \textbf{3.1640e-2 (7.55e-5)} \\
    RWMOP22 & 80    & 2     & 9     & 20000 & 0.0000e+0 (0.00e+0) & \textbf{8.2955e-1 (2.27e-1)} \\
    RWMOP23 & 80    & 2     & 6     & 20000 &0.0000e+0 (0.00e+0) & \textbf{9.9856e-1 (4.52e-16)} \\
    RWMOP24 & 105   & 3     & 9     & 26250 & 0.0000e+0 (0.00e+0) & 0.0000e+0 (0.00e+0) \\
    RWMOP25 & 80    & 2     & 2     & 20000 & \textbf{2.4079e-1 (1.79e-4) =} & 2.4078e-1 (1.83e-4) \\
    \midrule
    \multicolumn{5}{c}{+/-/=}             & 6/13/3 &  \\
    \bottomrule
    \end{tabular}%
    \end{adjustbox}
  \label{tab:addlabel}%
\end{table}%
\section{Comparison of the runtime of the proposed algorithm with other top-performing algorithms}
\label{sec:sample:runtime}

% Table generated by Excel2LaTeX from sheet 'runtime'
\begin{table}[H]
  \centering
  \color{black}
  \caption{Average runtime and Wilcoxon signed rank test of $I^{c}_{SDE^+}$ compared with the top performing CMOEAs on CF test suite over 30 independent runs }  
  \begin{adjustbox}{width=0.55\textwidth,center}
    \begin{tabular}{cccc}
    \toprule
    Problem & CCMO  & MOEADDAE &  $I^{c}_{SDE^+}$ \\
    \midrule
    CF1   & 5.3806e+1 (1.33e+1) = & 1.4849e+3 (1.78e+3) - & \textbf{4.7294e+1 (2.43e+0)} \\
    CF2   & 5.6501e+1 (1.34e+1) - & 1.4613e+3 (1.82e+3) - & \textbf{4.6827e+1 (2.41e+0)} \\
    CF3   & 5.5423e+1 (1.53e+1) = & 1.6043e+3 (1.99e+3) - & \textbf{4.6519e+1 (2.51e+0)} \\
    CF4   & 5.3079e+1 (1.49e+1) = & 1.5136e+3 (1.86e+3) - & \textbf{4.3246e+1 (2.20e+0)} \\
    CF5   & 5.8269e+1 (1.56e+1) = & 1.6639e+3 (2.00e+3) - & \textbf{4.9272e+1 (3.20e+0)} \\
    CF6   & 6.2093e+1 (1.50e+1) - & 1.5896e+3 (1.85e+3) - & \textbf{4.6212e+1 (2.61e+0)} \\
    CF7   & 5.6060e+1 (1.60e+1) - & 1.6909e+3 (2.04e+3) - & \textbf{4.6043e+1 (2.95e+0)} \\
    CF8   & 1.1473e+2 (5.55e+1) - & 1.4112e+3 (1.37e+3) - & \textbf{5.4319e+1 (3.05e+0)} \\
    CF9   & 1.2295e+2 (4.99e+1) - & 1.5200e+3 (1.50e+3) - & \textbf{5.6188e+1 (3.14e+0)} \\
    CF10  & 1.4915e+2 (6.18e+1) - & 1.3621e+3 (1.40e+3) - & \textbf{5.6284e+1 (3.36e+0)} \\
    \midrule
    +/-/= & 0/6/4 & 0/10/0 &  \\
    \bottomrule
    \end{tabular}%
    \end{adjustbox}
  \label{tab:addlabel}%
\end{table}%

% Table generated by Excel2LaTeX from sheet 'runtime'
\begin{table}[H]
  \centering
  \color{black}
  \caption{Average runtime and Wilcoxon signed rank test of $I^{c}_{SDE^+}$ compared with the top performing CMOEAs on CDTLZ test suite over 30 independent runs}  
  \begin{adjustbox}{width=0.55\textwidth,center}
    \begin{tabular}{cccc}
    \toprule
    Problem & CCMO  & MOEADDAE &  $I^{c}_{SDE^+}$ \\
    \midrule
    C1\_DTLZ1 & 1.1161e+1 (6.79e-1) - & 3.1398e+1 (1.57e+0) - & \textbf{7.6081e+0 (6.51e-1)} \\
    C1\_DTLZ3 & 2.2287e+1 (1.09e+0) - & 6.1902e+1 (1.21e+0) - & \textbf{1.5655e+1 (8.20e-1)} \\
    C2\_DTLZ2 & 8.8519e+0 (3.83e-1) - & 1.5731e+1 (4.87e-1) - & \textbf{3.7669e+0 (2.44e-1)} \\
    C3\_DTLZ1 & 1.2233e+1 (7.58e-1) - & 4.6348e+1 (1.44e+0) - & \textbf{1.0961e+1 (7.38e-1)} \\
    C3\_DTLZ4 & 1.1961e+1 (5.97e-1) - & 4.3913e+1 (2.50e+0) - & \textbf{1.1155e+1 (6.49e-1)} \\
    \midrule
    +/-/= & 0/5/0 & 0/5/0 &  \\
    \bottomrule
    \end{tabular}%
    \end{adjustbox}
  \label{tab:addlabel}%
\end{table}%

% Table generated by Excel2LaTeX from sheet 'runtime'
\begin{table}[H]
  \centering
  \color{black}
  \caption{Average runtime and Wilcoxon signed rank test of $I^{c}_{SDE^+}$ compared with the top performing CMOEAs on DASCMOP test suite over 30 independent runs}  
  \begin{adjustbox}{width=0.55\textwidth,center}
    \begin{tabular}{cccc}
    \toprule
    Problem & CCMO  & MOEADDAE &  $I^{c}_{SDE^+}$ \\
    \midrule
    DASCMOP1 & 1.2636e+2 (2.53e+1) - & 3.2581e+2 (3.84e+1) - & \textbf{6.8024e+1 (3.26e+0)} \\
    DASCMOP2 & 1.2554e+2 (2.34e+1) - & 3.0924e+2 (5.00e+1) - & \textbf{7.0081e+1 (3.77e+0)} \\
    DASCMOP3 & 1.3597e+2 (2.80e+1) - & 3.4440e+2 (3.09e+1) - & \textbf{6.8573e+1 (3.81e+0)} \\
    DASCMOP4 & 1.4281e+2 (2.54e+1) - & 3.7052e+2 (3.39e+1) - & \textbf{6.7401e+1 (3.85e+0)} \\
    DASCMOP5 & 1.4286e+2 (2.41e+1) - & 3.8862e+2 (3.52e+1) - & \textbf{6.7428e+1 (3.50e+0)} \\
    DASCMOP6 & 1.3750e+2 (2.00e+1) - & 3.7109e+2 (3.67e+1) - & \textbf{6.7464e+1 (3.71e+0)} \\
    DASCMOP7 & 1.6368e+2 (2.15e+1) - & 1.0958e+3 (1.75e+2) - & \textbf{6.9707e+1 (3.56e+0)} \\
    DASCMOP8 & 1.7232e+2 (2.74e+1) - & 4.0720e+2 (5.16e+1) - & \textbf{6.9436e+1 (3.70e+0)} \\
    DASCMOP9 & 1.5752e+2 (2.64e+1) - & 3.6717e+2 (4.31e+1) - & \textbf{7.1589e+1 (3.96e+0)} \\
    \midrule
    +/-/= & 0/9/0 & 0/9/0 &  \\
    \bottomrule
    \end{tabular}%
    \end{adjustbox}
  \label{tab:addlabel}%
\end{table}%

% Table generated by Excel2LaTeX from sheet 'runtime'
\begin{table}[H]
  \centering
  \color{black}
  \caption{Average runtime and Wilcoxon signed rank test of $I^{c}_{SDE^+}$ compared with the top performing CMOEAs on LIRCMOP test suite over 30 independent runs}  
  \begin{adjustbox}{width=0.55\textwidth,center}
    \begin{tabular}{cccc}
    \toprule
    Problem & CCMO  & MOEADDAE &  $I^{c}_{SDE^+}$ \\
    \midrule
    LIRCMOP1 & 9.8285e+1 (2.29e+1) - & 7.8887e+2 (2.14e+2) - & \textbf{6.5906e+1 (2.40e+0)} \\
    LIRCMOP2 & 1.0335e+2 (2.38e+1) - & 7.9192e+2 (2.23e+2) - & \textbf{6.6497e+1 (3.73e+0)} \\
    LIRCMOP3 & 1.0321e+2 (2.39e+1) - & 7.9209e+2 (2.29e+2) - & \textbf{6.6325e+1 (3.62e+0)} \\
    LIRCMOP4 & 1.0084e+2 (2.45e+1) - & 8.3836e+2 (2.32e+2) - & \textbf{6.6307e+1 (3.59e+0)} \\
    LIRCMOP5 & 1.1738e+2 (2.74e+1) - & 8.2083e+2 (2.36e+2) - & \textbf{7.6849e+1 (3.53e+0)} \\
    LIRCMOP6 & 1.2270e+2 (3.26e+1) - & 8.0664e+2 (2.38e+2) - & \textbf{7.1781e+1 (4.06e+0)} \\
    LIRCMOP7 & 1.1131e+2 (2.99e+1) - & 8.0885e+2 (2.38e+2) - & \textbf{6.8823e+1 (3.80e+0)} \\
    LIRCMOP8 & 1.0693e+2 (2.78e+1) - & 8.0203e+2 (2.39e+2) - & \textbf{6.8778e+1 (3.88e+0)} \\
    LIRCMOP9 & 1.3226e+2 (3.64e+1) - & 8.5781e+2 (2.39e+2) - & \textbf{6.8364e+1 (3.71e+0)} \\
    LIRCMOP10 & 1.7608e+2 (3.57e+1) - & 8.9821e+2 (2.46e+2) - & \textbf{7.1013e+1 (4.18e+0)} \\
    LIRCMOP11 & 2.4547e+2 (6.05e+1) - & 8.7527e+2 (2.31e+2) - & \textbf{6.6261e+1 (3.28e+0)} \\
    LIRCMOP12 & 1.1268e+2 (2.84e+1) - & 8.7629e+2 (2.42e+2) - & \textbf{6.6909e+1 (3.72e+0)} \\
    LIRCMOP13 & 8.0374e+2 (1.96e+2) - & 1.0579e+3 (2.82e+2) - & \textbf{8.2172e+1 (4.66e+0)} \\
    LIRCMOP14 & 4.5919e+2 (1.14e+2) - & 8.2307e+2 (2.35e+2) - & \textbf{7.6356e+1 (4.27e+0)} \\
    \midrule
    +/-/= & 0/14/0 & 0/14/0 &  \\
    \bottomrule
    \end{tabular}%
    \end{adjustbox}
  \label{tab:addlabel}%
\end{table}%
\vspace{-0.5cm}
% Table generated by Excel2LaTeX from sheet 'runtime'
\begin{table}[h]
  \centering
  \color{black}
  \caption{Average runtime and Wilcoxon signed rank test of $I^{c}_{SDE^+}$ compared with the top performing CMOEAs on MW test suite over 30 independent runs}
    \begin{adjustbox}{width=0.55\textwidth,center}
    \begin{tabular}{cccc}
    \toprule
    Problem & CCMO  & MOEADDAE &  $I^{c}_{SDE^+}$ \\
    \midrule
    MW1   & 1.1566e+1 (7.13e-1) - & 3.5409e+1 (2.30e+0) - & \textbf{9.4391e+0 (5.59e-1)} \\
    MW2   & 1.7533e+1 (5.74e-1) - & 4.1514e+1 (2.01e+0) - & \textbf{9.5179e+0 (5.72e-1)} \\
    MW3   & 9.7101e+0 (2.65e-1) - & 4.2489e+1 (1.39e+0) - & \textbf{9.0432e+0 (4.05e-1)} \\
    MW4   & 1.6895e+1 (6.58e-1) - & 3.8633e+1 (1.65e+0) - & \textbf{1.0691e+1 (6.40e-1)} \\
    MW5   & \textbf{8.5062e+0 (7.65e-1) +} & 3.8057e+1 (1.77e+0) - & 9.6911e+0 (5.68e-1) \\
    MW6   & 1.5495e+1 (5.83e-1) - & 4.1618e+1 (2.26e+0) - & \textbf{9.4568e+0 (4.88e-1)} \\
    MW7   & \textbf{7.5061e+0 (4.11e-1) +} & 4.2582e+1 (1.63e+0) - & 9.1033e+0 (5.25e-1) \\
    MW8   & 2.2817e+1 (9.10e-1) - & 4.2741e+1 (1.70e+0) - & \textbf{1.0024e+1 (5.73e-1)} \\
    MW9   & \textbf{7.7217e+0 (9.58e-1) +} & 4.0111e+1 (2.82e+0) - & 9.5472e+0 (5.87e-1) \\
    MW10  & 1.2889e+1 (1.99e+0) - & 4.1001e+1 (3.49e+0) - & \textbf{9.3998e+0 (5.57e-1)} \\
    MW11  & \textbf{6.8012e+0 (3.07e-1) +} & 4.1485e+1 (2.27e+0) - & 9.1270e+0 (5.61e-1) \\
    MW12  & \textbf{8.0080e+0 (1.40e+0) +} & 4.0589e+1 (2.25e+0) - & 9.7455e+0 (5.86e-1) \\
    MW13  & 1.1397e+1 (5.22e-1) - & 4.2332e+1 (1.59e+0) - & \textbf{9.4123e+0 (5.31e-1)} \\
    MW14  & 1.9929e+1 (5.80e-1) - & 4.1496e+1 (1.13e+0) - & \textbf{1.0624e+1 (6.29e-1)} \\
    \midrule
    +/-/= & 5/9/0 & 0/14/0 &  \\
    \bottomrule
    \end{tabular}%
    \end{adjustbox}
  \label{tab:addlabel}%
\end{table}%
\vspace{-0.5cm}
% Table generated by Excel2LaTeX from sheet 'runtime'
\begin{table}[H]
  \centering
  \color{black}
  \caption{Average runtime and Wilcoxon signed rank test of $I^{c}_{SDE^+}$ compared with the top performing CMOEAs on RWMOP test suite over 30 independent runs}  
  \begin{adjustbox}{width=0.55\textwidth,center}
    \begin{tabular}{cccc}
    \toprule
    Problem & CCMO  & MOEADDAE &  $I^{c}_{SDE^+}$ \\
    \midrule
    RWMOP1 & 3.3013e+0 (2.79e-1) - & 8.1471e+0 (7.54e-1) - & \textbf{3.1263e+0 (3.34e-1)} \\
    RWMOP2 & \textbf{2.4775e+0 (1.99e-1) +} & 8.6168e+0 (7.03e-1) - & 3.0181e+0 (3.32e-1) \\
    RWMOP3 & 4.1495e+0 (2.62e-1) - & 7.8110e+0 (6.68e-1) - & \textbf{2.8638e+0 (2.79e-1)} \\
    RWMOP4 & 4.0741e+0 (4.73e-1) - & 7.7945e+0 (6.17e-1) - & \textbf{2.9666e+0 (3.04e-1)} \\
    RWMOP5 & 3.4746e+0 (2.12e-1) - & 7.9449e+0 (6.32e-1) - & \textbf{2.9563e+0 (2.95e-1)} \\
    RWMOP6 & \textbf{2.5144e+0 (1.69e-1) +} & 7.2714e+0 (5.29e-1) - & 2.7822e+0 (2.53e-1) \\
    RWMOP7 & 3.0358e+0 (2.33e-1) = & 1.0849e+1 (1.41e+0) - & \textbf{2.9778e+0 (2.53e-1)} \\
    RWMOP8 & 1.3504e+1 (7.12e-1) - & 1.1562e+1 (9.76e-1) - & \textbf{3.8850e+0 (2.61e-1)} \\
    RWMOP9 & 5.2468e+0 (2.92e-1) - & 9.4424e+0 (7.74e-1) - & \textbf{3.0347e+0 (2.22e-1)} \\
    RWMOP10 & 7.5630e+0 (3.12e-1) - & 8.4465e+0 (5.66e-1) - & \textbf{2.9911e+0 (2.33e-1)} \\
    RWMOP11 & 1.3263e+2 (5.10e+0) - & 6.1581e+1 (4.71e+0) - & \textbf{1.0954e+1 (7.26e-1)} \\
    RWMOP12 & 6.5186e+0 (3.06e-1) - & 7.8349e+0 (5.23e-1) - & \textbf{2.6374e+0 (2.05e-1)} \\
    RWMOP13 & 7.5456e+0 (3.36e-1) - & 1.2744e+1 (7.84e-1) - & \textbf{3.9127e+0 (3.01e-1)} \\
    RWMOP14 & 3.4608e+0 (2.06e-1) - & 7.6157e+0 (5.11e-1) - & \textbf{2.9329e+0 (2.24e-1)} \\
    RWMOP15 & \textbf{2.4360e+0 (1.21e-1) +} & 8.1856e+0 (5.05e-1) - & 2.8188e+0 (1.80e-1) \\
    RWMOP16 & 7.8809e+0 (3.28e-1) - & 7.4304e+0 (5.10e-1) - & \textbf{2.6762e+0 (1.92e-1)} \\
    RWMOP17 & 4.1276e+0 (2.86e-1) - & 1.1490e+1 (9.00e-1) - & \textbf{3.9230e+0 (2.58e-1)} \\
    RWMOP18 & 6.3650e+0 (2.89e-1) - & 8.6638e+0 (5.14e-1) - & \textbf{2.9546e+0 (1.84e-1)} \\
    RWMOP19 & \textbf{3.8214e+0 (3.12e-1) +} & 1.1364e+1 (1.01e+0) - & 4.1341e+0 (2.48e-1) \\
    RWMOP20 & \textbf{2.2086e+0 (1.57e-1) +} & 7.1607e+0 (4.89e-1) - & 3.8292e+0 (3.51e-1) \\
    RWMOP21 & 6.3271e+0 (2.93e-1) - & 8.5191e+0 (5.68e-1) - & \textbf{2.9949e+0 (1.78e-1)} \\
    RWMOP22 & \textbf{1.7339e+0 (1.52e-1) +} & 6.4549e+0 (7.59e-1) - & 2.7836e+0 (7.19e-1) \\
    RWMOP23 & \textbf{2.1377e+0 (2.48e-1) +} & 8.0911e+0 (6.71e-1) - & 2.5790e+0 (1.98e-1) \\
    RWMOP24 & 3.8449e+0 (2.66e-1) - & 9.6932e+0 (7.31e-1) - & \textbf{3.5342e+0 (3.11e-1)} \\
    RWMOP25 & 8.5015e+0 (4.93e-1) - & 8.4330e+0 (6.69e-1) - & \textbf{2.7659e+0 (2.87e-1)} \\
    \midrule
    +/-/= & 7/17/1 & 0/25/0 &  \\
    \bottomrule
    \end{tabular}%
    \end{adjustbox}
  \label{tab:addlabel}%
\end{table}%

\vspace{30pt}
\printcredits

\bibliographystyle{cas-model2-names}

\bibliography{cas-refs}
\end{document}